\documentclass[journal]{IEEEtran}
\usepackage{amsmath,amsfonts}
\usepackage{array}
\usepackage[caption=false,font=normalsize,labelfont=sf,textfont=sf]{subfig}
\usepackage{textcomp}
\usepackage{stfloats}
\usepackage{url}
\usepackage{verbatim}
\usepackage{graphicx}
\usepackage{cite}
\usepackage{multirow}
\usepackage{hyperref}

\usepackage[linesnumbered,ruled,vlined]{algorithm2e}
\usepackage{algpseudocode} 
\usepackage{booktabs}
\usepackage{amsthm}
\usepackage{amssymb}
\usepackage[table]{xcolor}
\usepackage{color}

\begin{document}

\title{A Pluggable Common Sense-Enhanced Framework for Knowledge Graph Completion}

\author{Guanglin Niu, Bo Li, Siling Feng
\thanks{Guanglin Niu and Bo Li are with the School of Artificial Intelligence, Beihang University. Siling Feng is with the College of Information and Communication Engineering, Hainan University.}
}

\markboth{Journal of \LaTeX\ Class Files,~Vol.~14, No.~8, August~2021}%
{Shell \MakeLowercase{\textit{et al.}}: A Sample Article Using IEEEtran.cls for IEEE Journals}


\maketitle

\begin{abstract}

Knowledge graph completion (KGC) tasks aim to infer missing facts in a knowledge graph (KG) for many knowledge-intensive applications. However, existing embedding-based KGC approaches primarily rely on factual triples, potentially leading to outcomes inconsistent with common sense. Besides, generating explicit common sense is often impractical or costly for a KG. To address these challenges, we propose a pluggable common sense-enhanced KGC framework that incorporates both fact and common sense for KGC. This framework is adaptable to different KGs based on their entity concept richness and has the capability to automatically generate explicit or implicit common sense from factual triples. Furthermore, we introduce common sense-guided negative sampling and a coarse-to-fine inference approach for KGs with rich entity concepts. For KGs without concepts, we propose a dual scoring scheme involving a relation-aware concept embedding mechanism. Importantly, our approach can be integrated as a pluggable module for many knowledge graph embedding (KGE) models, facilitating joint common sense and fact-driven training and inference. The experiments illustrate that our framework exhibits good scalability and outperforms existing models across various KGC tasks.

\end{abstract}


\section{Introduction}

\IEEEPARstart{K}{nowledge} graphs (KGs) such as Freebase~\cite{BGF:Freebase}, YAGO~\cite{Yago}, WordNet~\cite{Miller:WordNet}, NELL~\cite{Mitchell:nell}, and DBpedia~\cite{Lehmann:dbpedia} store factual triples in the form of $(head\ entity,\ relation,$ $\ tail\ entity)$ (shortened as $(h, r, t)$), which can be applied to numerous knowledge-intensive tasks, including relation extraction\cite{REtask}, semantic search~\cite{searchtask}, dialogue systems~\cite{Dialogtask}, question answering~\cite{QAtask}, and recommender systems~\cite{rectask}. Particularly, many KGs contain ontologies that comprising multiple entity concepts along with their associated meta-relations.

The primary issue of employing KGs is their natural incompleteness, primarily attributed to noisy data and the limited performance of current information extraction models~\cite{KGCtnnls}. Knowledge graph completion (KGC) is a vital endeavor that aims to resolve this issue by inferring missing entities or relations in unobserved triples represented as $(h, r, ?)$ or $(?, r, t)$. To accomplish this, knowledge graph embedding (KGE) is a predominant technique to learn entity and relation embeddings that can be employed to assess the plausibility of unseen triple candidates through scoring mechanisms~\cite{KGEtkde}.

\begin{figure*}
    \centering
    \includegraphics[width=1.0\textwidth]{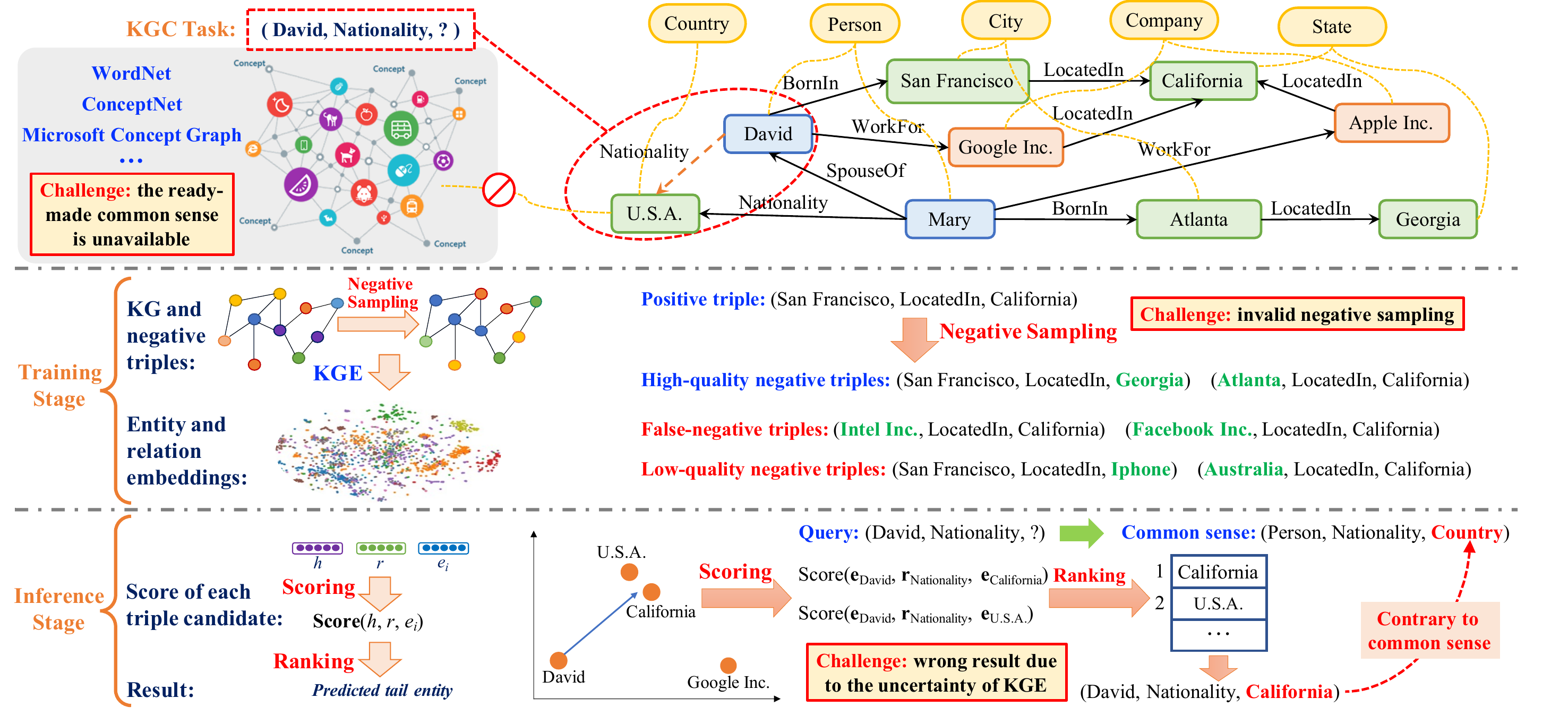}
    \caption{Illustration of the KGC task and three main challenges of the KGE technique. The top half shows a KG with triples and ontological concepts linked to entities as well as a KGC task of tail entity missing $(David, Nationality, ?)$. This figure highlights three challenges of KGE models on KGC tasks: (1) Lack of ready-made common sense that can be directly used for entity-specific KGC tasks. (2) Invalid negative sampling during the training procedure of a KGE model which would limit its performance. (3) Incorrect results of KGC due to the uncertainty of KGE at the inference stage.} 
    \label{fig:intro}
\end{figure*}

 TransE~\cite{Bordes:TransE} is a widely-used KGE approach that models relations as translation operations from the head to tail entities in a triple $(h, r, t)$, formulated as $\mathbf{h}+\mathbf{r}=\mathbf{t}$. To enhance the representation ability, many variants of TransE have been developed, including TransH~\cite{Wang:TransH}, TransR~\cite{Lin:TransR}, RotatE~\cite{RotatE}, QuatE~\cite{QuatE} and HAKE~\cite{HAKE}. Tensor factorization-based approaches like RESCAL~\cite{RESCAL} and DistMult~\cite{Distmult} utilize tensor products and diagonal matrices to capture complex interactions among latent factors. ComplEx~\cite{Trouillon:ComplEx} extends DistMult by representing entities and relations in the complex-valued space to model asymmetric relations effectively. Some KGE models exploit deep neural network to predict the plausibility of a triple, such as fully connected neural network-based model NTN~\cite{NTN}, convolutional neural network-based approach ConvE~\cite{Dettmers:ConvE}, graph neural network-based method R-GCN~\cite{RGCN}, transformer-based model KG-BERT~\cite{KG-BERT}.

During the inference stage, candidate triples are scored and sorted to produce inference results. \textbf{However, this process may rank incorrect entities higher than correct ones due to uncertainty of KG embeddings.} Some KGC techniques augment entity embeddings with external informantion, such as text descriptions~\cite{DKRL, TEKE} or images~\cite{Xie:IKRL, RSME}. Whereas, the extra multi-modal information is usually unavailable. Conversely, humans always utilize common sense to directly evaluate the plausibility of facts. For instance, in Fig.~\ref{fig:intro}, a KGE model believes the predicted tail entity $California$ as the highest-ranked candidate but the corresponding concept $State$ is inconsistent with the common sense $(Person, Nationality, Country)$. \textbf{Thus, taking advantage of common sense for KGC is a key idea of our work.}

In contrast to factual triples, common sense is commonly represented as concepts together with their relations in the format of $(head\ concept,$ $relation, tail \ concept)$ in some popular common sense KGs like ConceptNet~\cite{conceptnet} and Microsoft Concept Graph~\cite{MicrosoftCG}. \textbf{However, common sense is costly the existing commonsense KGs only contain concepts without links to corresponding entities, which is not applicable to entity-centric KGC tasks.} Although some KGE models leverage ontology for incorporating common sense-like information, such as TKRL~\cite{Xie:TKRL} with hierarchical entity types and JOIE~\cite{JOIE} introducing ontology layer embeddings, these ontology-based models cannot work on some KGs lacking entity concepts such as WordNet~\cite{Miller:WordNet}. \textbf{Therefore, generating common sense automatically from any KG remains a challenge of exploiting common sense for KGC.}

Following the open-world assumption~\cite{OWA}, most KGE models employ a pairwise loss function and employ negative sampling process to negerate negative triples based on the local closed-world assumption~\cite{Part-Close-World-Assumpt}. \textbf{However, uniform sampling~\cite{Wang:TransH} cannot judge the validity of negative triples and might generate low-quality or false-negative triples.} As the instance shown in Fig.~\ref{fig:intro}, the semantic gap between a low-quality negative triple $(San\ Francisco, LocatedIn, Iphone)$ and a positive triple $(San\ Francisco, LocatedIn, California)$ is too large, causing invalid training of KGE models. Although some mechanisms like KBGAN~\cite{kbgan} and self-adversarial sampling~\cite{RotatE} evaluate the quality of negative triples, the issue of false negatives persists. \textbf{Therefore, generating high-quality triples is crucial for training any KGE model effectively}.

\begin{figure*}
    \centering
    \includegraphics[width=1.0\textwidth]{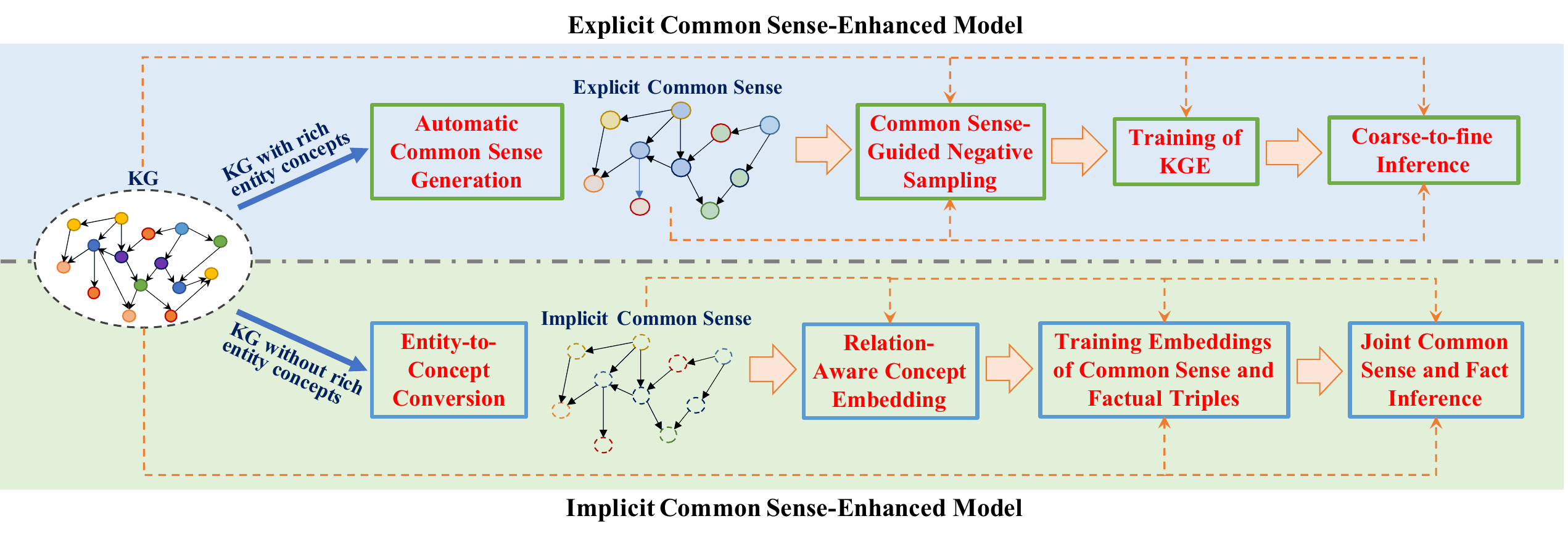}
    \caption{The brief structure of our proposed framework. The upper part is the explicit common sense-enhanced model, containing a simple yet effective automatic common sense generation module that could produce valuable explicit common sense and facilitate common sense-guided negative sampling as well as coarse-to-fine inference. Contrarily, the lower part exhibits the implicit common sense-enhanced model. Particularly, it introduces relation-aware concept embeddings for representing implicit common sense, and then conducts joint inference in the view of both common sense and fact.}
    \label{fig:framework}
\end{figure*}

To address the above pivotal challenges, a pluggable common sense-enhanced KGC framework is proposed. As illustrated in Fig.~\ref{fig:framework}, our framework consists of an \underline{\bf{E}}xplicit \underline{\bf{C}}ommon \underline{\bf{S}}ense-\underline{\bf{E}}nhanced (ECSE) model (in section~\ref{sec:ECSE}) and an \underline{\bf{I}}mplicit \underline{\bf{C}}ommon \underline{\bf{S}}ense-\underline{\bf{E}}nhanced (ICSE) scheme (in section~\ref{sec:Auto}). Specifically, explicit common sense could be automatically generated with ontological concepts. On account of KGs lacking concepts, each factual triple could be extended to an implicit common sense-level triple. For instance, given a factual triple $(David,$ $WorkFor,$ $Google\ Inc.)$, we could generate an explicit common sense triple $(Person,$ $WorkFor,$ $Company)$, or an implicit one $(David's$ $concept,$ $WorkFor,$ $ Google$ $Inc.'s$ $concept)$ in the absence of entity concepts. Based on explicit common sense, a common sense-guided high-quality negative sampling strategy is designed to construct high-quality negative triples to facilitate more effective training of KGE models. Furthermore, a novel coarse-to-fine inference mechanism is proposed to ensure that the predicted triples conform to common sense. On the other hand, we develop a relation-aware concept embedding mechanism to learn the representation of implicit common sense triples, and then score each candidate triple based on both the common sense-level and the factual triples.

The main contributions of this paper can be summarized as the following three-folds:
\begin{itemize}
    \item To the best of our knowledge, it is the first effort to introduce common sense into KGE in both training and inference stages, contributing to higher accuracy of KGC in a joint common sense and fact-driven fashion. More interestingly, our framework can be conveniently integrated as a pluggable module into existing KGE models.
    
    \item We rigorously demonstrate that our ICSE model could represent both factual and common sense-level triples across various relation patterns and complex relations.

    \item To evaluate the scalability and effectiveness of our proposed framework, we conduct extensive experiments compared with some typical KGC baseline models and negative sampling strategies in the scenarios with and without entity concepts to demonstrate the superiority of the proposed framework.
\end{itemize}


\section{Related Works}

\subsection{Classical KGE Models}

In comparison to rule learning-based models~\cite{Galarrage:AMIE, RPJE, AnyBurl} and multi-hop reasoning-based models~\cite{DeepPath, Lin:reward-shaping, pathneurocomputing}, KGE approaches demonstrate superior efficiency, robustness, and inference performance. At present, the representative KGE models could be classified into three main categories:

(1) Translation-based models. One of the most typical KGE models TransE~\cite{Bordes:TransE} regards the relation in a triple as the translation operation from the head to the tail entities. TransH~\cite{Wang:TransH} and TransR~\cite{Lin:TransR} extend TransE by defining a hyperplane and a space specific to each relation, addressing the issue of inferring complex relations namely one-to-many (1-N), many-to-one (N-1), and many-to-many (N-N). RotatE~\cite{RotatE} represents symmetric relations by regarding the relation as a rotation operation. QuatE~\cite{QuatE} represents entities and relations in quaternion space to enhance the representation. HAKE~\cite{HAKE} learns entity and relation embeddings in the polar coordinate system to represent entities at different hierarchy levels.

(2) Tensor factorization-based models. RESCAL~\cite{RESCAL} score each triple via three-way matrix multiplication among two vectors of the entity pair and a matrix representing the relation. DistMult~\cite{Distmult} simplifies RESCAL by representing each relation with a diagonal matrix. ComplEx~\cite{Trouillon:ComplEx} embeds entities and relations into complex space, and performs tensor decomposition with Hamiltonian multiplication. HolE~\cite{HolE} models the interaction between entities via vector circular correlation. DURA~\cite{DURA} designs an entity embedding regularizer for tensor factorization-based models.

(3) Neural network-based models. NTN~\cite{NTN} and NAM~\cite{NAM} employ multi-layer perception while ConvE~\cite{Dettmers:ConvE} and ConvKB~\cite{ConvKB} exploit convolutional neural networks to encode interactions among entities and relations. R-GCN~\cite{RGCN}, SACN~\cite{SACN}, KBGAT~\cite{KBGAT} and DRGI~\cite{DRGI} introduce graph neural networks to encode the neighborhood of entities.

\subsection{KGE Models Based on Auxiliary Information}

Conventional KGE models exclusively focus on factual triples within KGs, neglecting a substantial amount of auxiliary information associated with entities and relations. TKRL~\cite{Xie:TKRL} utilizes entity types but might introduce some noisy types. JOIE~\cite{JOIE} jointly learns the embeddings from both the ontology and the instance graphs. Nevertheless, the ontology is inapplicable to inference because the relations in the ontology have few overlaps with those in the instance graph. Moreover, many KGs such as NELL~\cite{Mitchell:nell}, constructed through automatic or semi-automatic OpenIE techniques~\cite{OpenIE}, only express the abstract concepts of entities. Additionally, some KGs like WordNet~\cite{Miller:WordNet} even lack entity types, which limits the effectiveness of current type-based KGC models.

DKRL~\cite{DKRL} and TEKE~\cite{TEKE} enhance the representation of entities by encoding texts in the entity embedding space. KG-BERT~\cite{KG-BERT} utilizes a BERT module~\cite{BERT} to encode the textual description associated with entities and relations, and then evaluates the plausibility of triples. IKRL~\cite{Xie:IKRL} focuses on fusing the visual features and the structural embeddings of entities. MKGformer~\cite{MKGformer} integrates visual and textual entity representations through a multi-level fusion module. However, it is important to note that the multi-modal information is always unavailable for KGs.

\subsection{Negative Sampling on KGs}

In accordance with the open-world assumption (OWA)~\cite{OWA}, training KGE models consistently employs a pair-wise loss function with both positive and negative triples. Existing negative sampling strategies for KGE models are designed based on the local-closed world assumption~\cite{KGEeval} and can be categorized into five groups: (1) Random uniform sampling: randomly substituting an entity or relation in the positive triple with another entity or relation, following a uniform distribution~\cite{Wang:TransH}. (2) Adversarial-based sampling: KBGAN~\cite{kbgan} learns KG embeddings within an adversarial training framework~\cite{GAN}. This allows the discriminator to select high-quality negative triples. Inspired by KBGAN, Self-adversarial sampling~\cite{RotatE} efficiently evaluates the quality of negative triples without a generator. (3) Domain constraint-based sampling: these negative sampling techniques corrupt entities according to the constraints derived from abstract domains~\cite{typeconstraint} or concrete type information~\cite{DomainSampling}. (4) Efficient sampling: NSCaching~\cite{zhang2019nscaching} incorporates a caching mechanism to efficiently sample negative triple candidates. (5) None-sampling: NS-KGE~\cite{nonesampling} eliminates negative triples by transforming the pair-wise loss function into a square loss.

\begin{figure}
    \centering
    \includegraphics[scale=0.37]{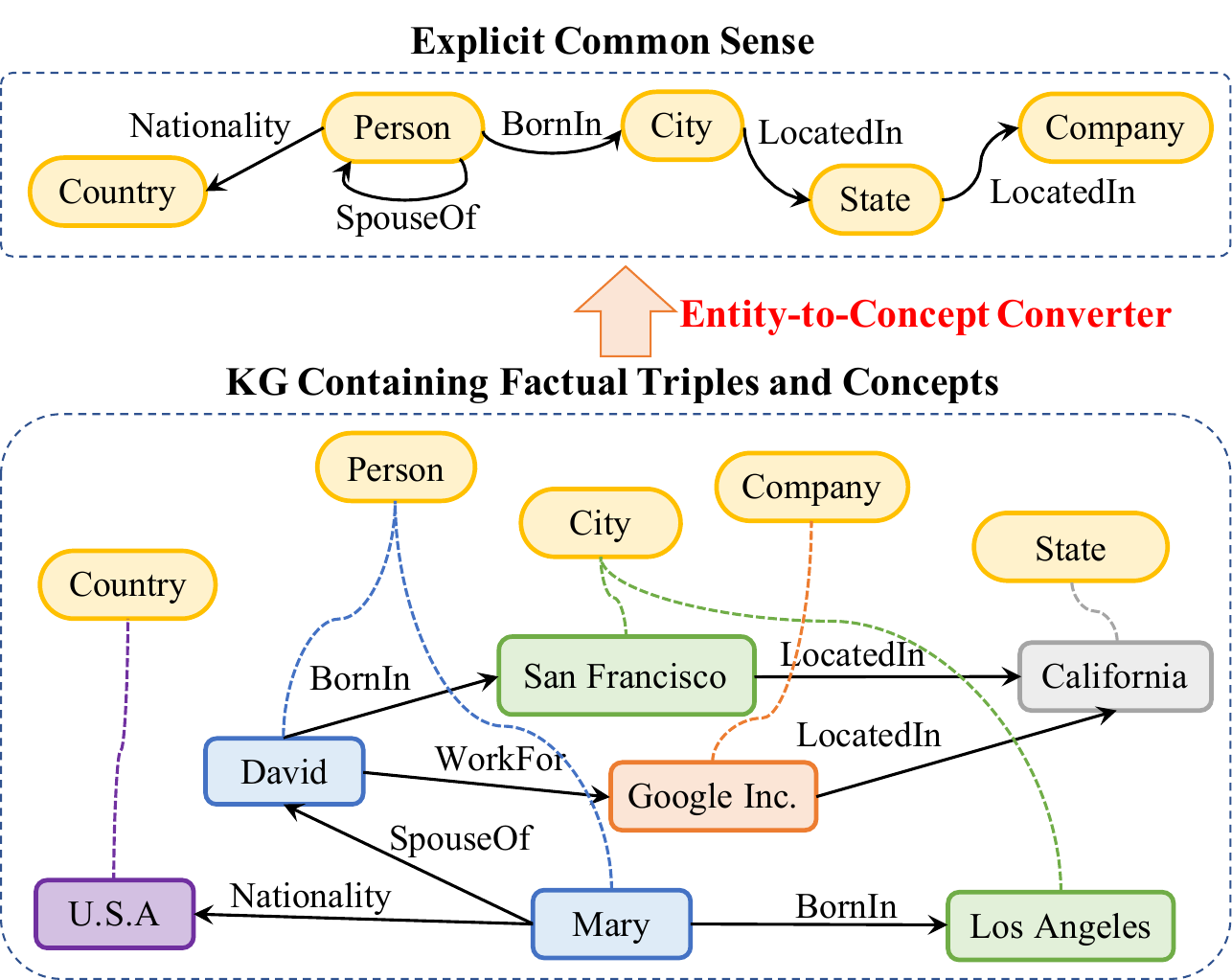}
    \caption{Illustration of the common sense generation module. Each entity is linked by a corresponding concept (in yellow) in the KG. Meanwhile, entities belonging to the same concept are shown in the same color. Then, the explicit common sense containing concepts with their relations could be generated and derived from the KG by an entity-to-concept converter.}
    \label{fig:common sense}
\end{figure}

\section{Explicit Common Sense-Guided Model}
\label{sec:ECSE}

For KGs containing rich entity concepts, explicit common sense can be automatically generated from the KG. Then, a common sense-guided negative sampling strategy is designed to improve the training process. Furthermore, we propose a coarse-to-fine inference mechanism that incorporates common sense-based candidate filtering and fact-based prediction. 

\subsection{Automatic Common Sense Generation}

Inspired by some well-known common sense graphs such as ConceptNet~\cite{conceptnet} and Microsoft Concept Graph~\cite{MicrosoftCG} which represent common sense as concepts linked by their relations, we generate appropriate common sense by substituting entities in factual triples with concepts via an entity-to-concept converter as shown in Fig.~\ref{fig:common sense}. Particularly, according to the requirements for the usage of common sense in the following negative sampling and inference procedures, the automatically generated common sense here can be separated into the individual-form $\mathcal{C}_1$ and the set-form $\mathcal{C}_2$ as:
\begin{equation}
    \mathcal{C}_1=\left(c_h,r,c_t\right), \ \ \ \mathcal{C}_2=\left(C_h,r,C_t\right) \label{eq1}
\end{equation}
where the individual-form common sense $\left(c_h,r,c_t\right)$ consists of a head concept $c_h$, a tail concept $c_t$ and an instance-level relation $r$ between them. Furthermore, the set-form of common sense $\mathcal{C}_2$ is derived by merging the common sense triples with the same relation in $\mathcal{C}_1$ into a single common sense triple consisting of a relation and the accompanying head concept set and tail concept set.

For better understanding, take the instance in Fig.~\ref{fig:common sense}, the common sense of individual-form associated with the relation $LocatedIn$ could be represented as $(City, LocatedIn, State)$ and $(Company, LocatedIn, State)$ while the common sense of set-form is $(\left\{City, Company\right\}, LocatedIn, \{State\})$. It is noteworthy that the individual-form is more accurate for representing common sense. On the contrary, the set-form common sense has a more diverse representation since the head concept and the tail concept are not unique. The detailed workflow of our automatic common sense generation mechanism is provided in Algorithm~\ref{alg1}.

\begin{algorithm}  
  \caption{The workflow of automatic common sense generation mechanism.}  
  \label{alg1}
  \noindent \KwIn{
$\mathcal{F}$: factual triples in the KG \newline
$e2c$: a dictionary converting each entity into its corresponding concept set
}
\KwOut{
$\mathcal{C}_1$: individual-form of common sense \newline
$\mathcal{C}_2$: set-form of common sense
}  
$C_1 = list()$; \tcp{Store $\mathcal{C}_1$ via a list}
$C_2 = dict()$; \tcp{Store $\mathcal{C}_2$ via a dictionary}
\For{$f\in \mathcal{F}$}{
  $f=(h, r, t)$\;
  $ch = e2c[h]$\;
  $ct = e2c[t]$\;
  \For{$ch_i\in ch$}
  {
    \For{$ct_i\in ct$}
    {
      $c_i=(ch_i, r, ct_i)$\;
      \If{$c_i$ not in $C_1$}
      {
        $C_1.append(c_i)$\;
      }
    }
  }
  \If{$r$ not in $C_2$}
  {
    $C_2[r] = [ch, ct]$\;
  } 
  \Else
  {
    $C_2[r] = [(ch \cup C_2[r][0]), (ct \cup C_2[r][1])]$\;
  }
}
\end{algorithm}

\subsection{Common Sense-Guided Negative Sampling}

Generating high-quality negative triples is a crucial aspect of training robust KGE models. To achieve this, it is essential to simultaneously address three key factors: (1) Preventing the erroneous negative triples which would introduce noise. (2) Acquiring negative triples of superior quality, thereby enhancing the overall robustness of the embedding model. (3) Establishing a diverse array of negative triples, contributing to the comprehensive evaluation and refinement of KGE models.


To address the challenge of avoiding false-negative triples while concurrently enhancing the quality and diversity of negative triples, we take advantage of the common sense in set-form and leverage the complex properties of relations. We modify the traditional negative sampling strategies by developing two sampling principles as followings.

Complex relation-aware sampling: this sampling strategy considers the complex characteristics of relations like N-1 relation $BirthPlace$. We define the unique entity and the non-unique entity according to the complex properties of relations, such as a non-unique tail entity and a unique head entity associated with the N-1 relation "BirthPlace". By replacing the unique entity in a positive triple by any other entity, the reconstructed triple must be incorrect, ensuring that it is a true negative triple and resolving the issue of false-negative triples. On the contrary, a negative triple created by replacing non-unique entities requires extra criteria to evaluate its quality.

Common sense-enhanced sampling: unlike random sampling, this common sense-enhanced sampling approach creates negative triples that exhibit semantic similarity to positive triples, contributing to more performance gains of training KGE models. Specific to the negative triple candidates obtained by substituting the unique entity, the higher score indicates the higher quality, which could be assigned with larger weights. Besides, the higher-scored negative triples achieved by replacing the non-unique entity are more likely to be potential positive triples, and their weights should be lower to lessen the influence of training with false-negative triples.

\begin{figure}
    \centering
    \includegraphics[scale=0.33]{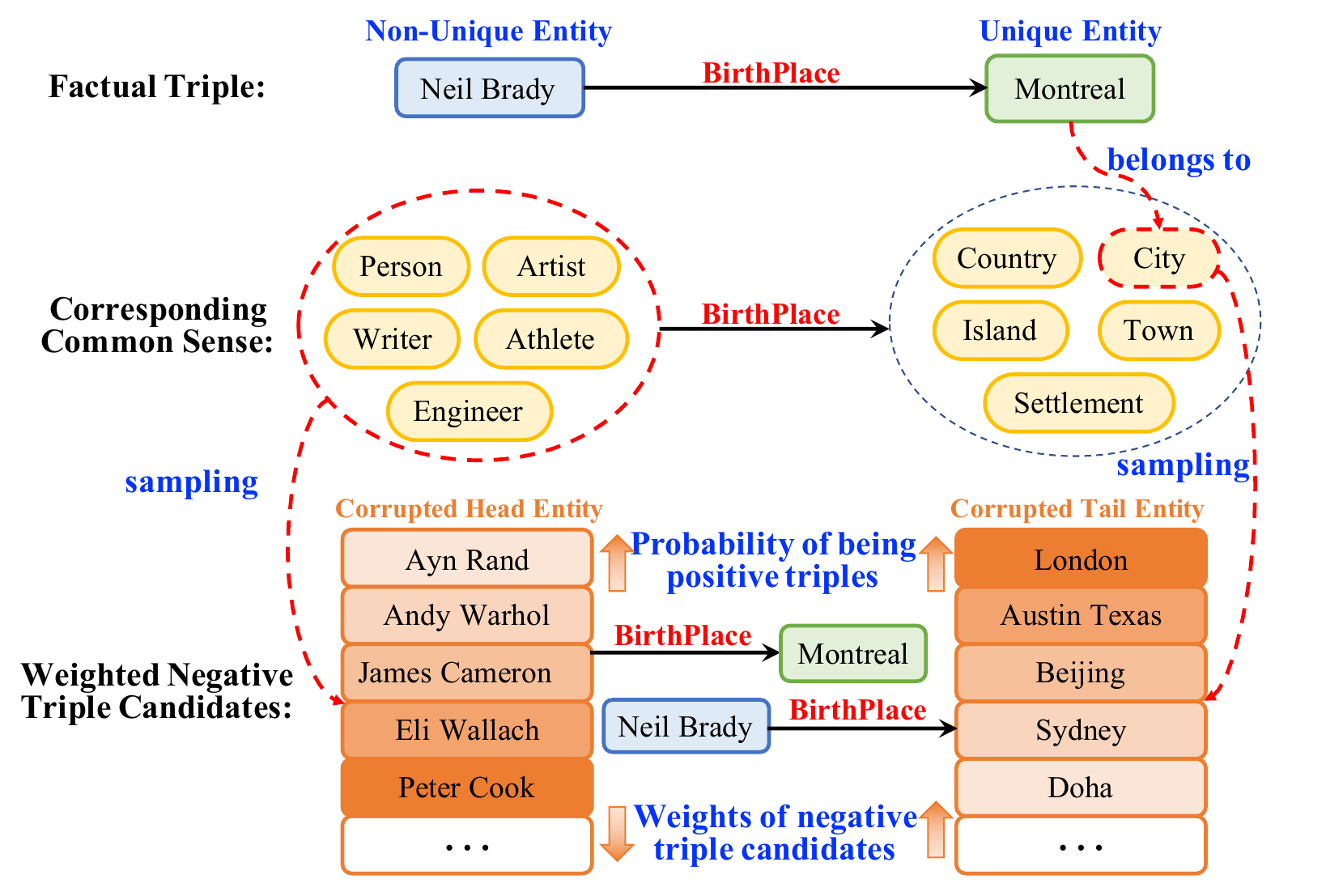}
    \caption{A case of common sense-guided negative sampling procedure specific to an N-1 relation $BirthPlace$. The common sense in set-form corresponding to the positive triple is presented. Besides, the arrow represents the direction of increasing value. The darker colors of negative triple candidates indicate larger weights for training KGE models.}
    \label{fig:negativesampling}
\end{figure}

To explain the advantages of our high-quality negative sampling, we present an example applied to an N-1 relation $BirthPlace$ on the dataset DBpedia as shown in Fig.~\ref{fig:negativesampling}. The negative sampling process comprises two main steps:

(1) Selecting concept candidates: given a positive triple $(Neil\ Brady, BirthPlace, Montreal)$, we extract the corresponding set-form common sense $(\{Person, Artist, Writer,$ $ Athlete, Engineer\}, BirthPlace, \{Country, City, Island,$ $Settlement, Town\})$ in regard to the relation $BirthPlace$. On account of the complex relation-aware sampling and the non-unique head entity, we identify all head concepts within the selected common sense as the head concept candidates. Besides, the concept $City$ is selected as the tail concept candidate with regard to the unique entity $Montreal$.

(2) Calculating the weights of negative triples: we sample entities belonging to the head or tail concept candidates to construct negative triple candidates such as $(Ayn\ Rand,$ $BirthPlace, Montreal)$. Then, we calculate the score of each negative triple candidate by KGE score function following the common sense-enhanced sampling principle. Specifically, the weight of a negative triple obtained by replacing the head entity $Neil\_Brady$ should be higher when the score of this negative triple is lower, thereby mitigating the false-negative issue. On the contrary, the weight of a negative triple generated by corrupting the tail entity $Montreal$ is higher with the higher score, ensuring the better quality of the negative triple.

Particularly, our framework is model-agnostic as to KGE, so we define a unified notation of score function $E_{ec}\left(h,r,t\right)$ to assess the plausibility of a triple $\left(h,r,t\right)$. Here are three most typical score functions utilized in KGE models:

(1) Translation-based score function, such as TransE:
\begin{equation}
    E_{ec}\left(h,r,t\right)=-\Vert \mathbf{h} + \mathbf{r} - \mathbf{t} \Vert \label{eq3}
\end{equation}
where \textbf{h}, \textbf{r} and \textbf{t} signify the vector embeddings of head entity $h$, relation $r$ and tail entity $t$, respectively.

(2) Rotation-based score function, such as RotatE:
\begin{equation}
    E_{ec}\left(h,r,t\right)=-\Vert \mathbf{h} \circ \mathbf{r} - \mathbf{t} \Vert \label{eq4}
\end{equation}
in which $\circ$ denotes Hardmard product, and $\mathbf{h}$, $\mathbf{r}$, $\mathbf{t}$ indicate the vector embeddings in the complex space.

(3) Tensor decomposition-based score function of ComplEx:
\begin{equation}
    E_{ec}(h,r,t)={\rm Re}(\mathbf{h}^\top {\rm diag}\left(\mathbf{r}\right)\overline{\mathbf{t}}) \label{eq5}
\end{equation}
where ${\rm diag}(\mathbf{r})$ is a diagonal matrix in the complex space corresponding to $r$. Besides, $\mathbf{h}$ and $\mathbf{t}$ are the complex vectors of $h$ and $t$. $\overline{\mathbf{t}}$ denotes the conjugation of $\mathbf{t}$. Then, the weight of each reconstructed negative triple is obtained as followings:
\begin{align}
    w\left(h_j^\prime,r,t\right)&=1-p\left(h_j^\prime,r,t\right) =1-\frac{exp\left(E_{ec}\left(h_j^\prime,r,t\right)\right)}{\sum_{i} e x p\left(E_{ec}\left(h_i^\prime,r,t\right)\right)} \\
    w\left(h,r,t_j^\prime\right)&=p\left(h,r,t_j^\prime\right) =\frac{exp{\left(E_{ec}\left(h,r,t_j^\prime\right)\right)}}{\sum_{i} e x p\left(E_{ec}\left(h,r,t_i^\prime\right)\right)}
\end{align}
where $p\left(h_j^\prime,r,t\right)$ and $p\left(h,r,t_j^\prime\right)$ are the probability of the negative triples $\left(h_j^\prime,r,t\right)$ and $\left(h,r,t_j^\prime\right)$ being positive triples, respectively. $w\left(h_j^\prime,r,t\right)$ and $w\left(h,r,t_j^\prime\right)$ indicate the weights of these two negative triples for training procedure. 

We introduce a weighting scheme for negative triples originating from non-unique entity corruption, denoted as $1-p$, mitigating the influence of false-negative triples. Conversely, in the context of negative triples derived from the unique entity, negative triples characterized by larger $p$ values serve as indicators of higher quality that are assigned with higher weights. Thus, we allocate the highest weights to negative triple candidates involving the head entity $Peter\ Cook$ and the tail entity $London$ as shown in Fig.~\ref{fig:negativesampling}.

Similar to the above-mentioned common sense-guided negative sampling process for N-1 relations, the negative triples could also be generated for 1-1, 1-N, and N-N relations. Algorithm~\ref{alg2} shows the general procedure for our developed common sense-guided negative sampling strategy, adaptable to diverse properties of complex relations.

\begin{algorithm}
\renewcommand{\thealgocf}{2}
  \caption{The algorithm of common sense-guided negative sampling strategy.}  
  \label{alg2}
  \noindent \KwIn{
$\mathcal{F}$: factual triples in the KG \newline
$\mathcal{C}_2$: set-form common sense \newline
$e2c$: a dictionary converting each entity into its corresponding concept set
}
\KwOut{
$\mathcal{N}$: weighted high-quality negative triples \newline
}  

$N = list()$; \tcp{Negative triples}
\For{$f\in \mathcal{F}$}{
  $f=(h, r, t)$;  $cs = C_2[r]$\;
  \If{$r$ is a 1-1 relation}
  {
    $ch = Sample(e2c[h])$; $ct = Sample(e2c[t])$\;
    \For{$h_j^\prime \in E_h$}
    {
      $neg = (h_j^\prime, r, t)$;      $w = p\left(h_j^\prime,r,t\right)$\;
    }
    \For{$t_j^\prime \in E_t$}
    {
      $neg = (h, r, t_j^\prime)$;      $w = p\left(h,r,t_j^\prime\right)$\;
    }
  }
  \If{$r$ is a 1-N relation}
  {
    $ch = Sample(e2c[h])$;    $ct = Sample(cs[1])$\;
    \For{$h_j^\prime \in E_h$}
    {
      $neg = (h_j^\prime, r, t)$;      $w = p\left(h_j^\prime,r,t\right)$\;
    }
    \For{$t_j^\prime \in E_t$}
    {
      $neg = (h, r, t_j^\prime)$;      $w = 1 - p\left(h,r,t_j^\prime\right)$\;
    }
  }
  \If{$r$ is a N-1 relation}
  {
    $ch = Sample(cs[0])$;    $ct = Sample(e2c[t])$\; 
    \For{$h_j^\prime \in E_h$}
    {
      $neg = (h_j^\prime, r, t)$;      $w = 1 - p\left(h_j^\prime,r,t\right)$\;
    }
    \For{$t_j^\prime \in E_t$}
    {
      $neg = (h, r, t_j^\prime)$;      $w = p\left(h,r,t_j^\prime\right)$\;
    }
  }
  \If{$r$ is a N-N relation}
  {
    $ch = Sample(cs[0])$;    $ct = Sample(cs[1])$\;
    \For{$h_j^\prime \in E_h$}
    {
      $neg = (h_j^\prime, r, t)$;      $w = 1 - p\left(h_j^\prime,r,t\right)$\;
    }
    \For{$t_j^\prime \in E_t$}
    {
      $neg = (h, r, t_j^\prime)$;      $w = 1 - p\left(h,r,t_j^\prime\right)$\;
    }
  }
  $N.append([neg, w])$\;
}
return N\;
\end{algorithm}

\subsection{Training and Coarse-to-Fine Inference}

Furthermore, we feed positive triples and weighted high-quality negative triples into a KGE model to learn entity and relation embeddings. Corresponding to the various score functions introduced above, we present two typical loss functions:
\begin{align}
    &L=\sum_{i}^{n}\{max[0,\gamma-E_{ec}\left(h,r,t\right)+w(h_i^\prime,r,t)E_{ec}(h_i^\prime,r,t)] \nonumber \\
    & + max[0,\gamma-E_{ec}(h,r,t)+w(h,r,t_i^\prime)E_{ec}(h,r,t_i^\prime)]\} \label{eq8}
\end{align}
\begin{align}
    &L=-\log\sigma(\gamma+E_{ec}\left(h,r,t\right)) - \sum_{i}^{n}{\{w(h_i^\prime,r,t)\log\sigma(-E_{ec}} \nonumber \\
    &(h_i^\prime,r,t)-\gamma)+w(h,r,t_i^\prime)\log\sigma(-E_{ec}(h,r,t_i^\prime)-\gamma)\} \label{eq9}
\end{align}
in which $\gamma$ denotes the margin, $max[0, x]$ means the function used to output the larger value between 0 and x. $\sigma$ indicates the sigmoid function. $n$ is the negative sampling size.

\begin{algorithm}  
  \caption{Training and coarse-to-fine inference procedures of ECSE model.}
  \label{alg3}
  \noindent \KwIn{
$\mathcal{F}$: factual triples in the KG \newline
$\mathcal{N}$: weighted negative triples \newline
$\mathcal{C}_1$: individual-form common sense \newline
$e2c$: a dictionary converting each entity into its corresponding concept set
}
\KwOut{
$\mathcal{P}$: inference results \newline
}
\tcp{Training procedure}
\For{$f \in \mathcal{F}$}
{
  Select the corresponding common sense in $\mathcal{C}_1$ with regard to $f$\;
  Extract negative triples in $\mathcal{N}$ associated with $f$\;
  Calculate loss function $L$ such as Eq.~\ref{eq8}-Eq.\ref{eq9}\;
  Optimize loss function $L$ to access entity and relation embeddings\;
}
\tcp{Inference procedure}
Test instance: $ts = (h, r, ?)$\;
$ch = e2c[h]$\;
$ct=\{ct_i | (ch, r, ct_i)\}$ $\leftarrow$ $Sample(\mathcal{C}_1, ch, r)$; \tcp{Sampling concepts from the common sense related to the concept set $ch$ together with the relation $r$}
$e_{cand} \leftarrow$ $Sample(\mathcal{E}, ct)$; \tcp{Sampling candidate entities belonging to the concepts in $ct$}
\For{$e_j \in e_{cand}$}
{
  Calculate $E_{ec}(h, r, e_j)$ according to the score function that used in the training procedure\;
}
Rank all the candidate triples $(h, r, e_j)$ in descending order according to their scores\;
$\mathcal{P} \leftarrow$ top-ranked $\{(h, r, e_j)\}$\;
return $\mathcal{P}$\;
\end{algorithm}

Intuitively, the KGC results aligned with common sense are more likely to be correct. Thus, we propose a coarse-to-fine inference mechanism in the views of both common sense and factual triples. Specifically, in the coarse-grained concept filtering phase, when confronted with a query triple containing a missing entity, all entities consistent with the tenets of common sense are selected to form the pool of triple candidates. Taking the tail entity prediction of KGC task $\left(Mary, Nationality, ?\right)$ as an instance, the common sense $(Person, Nationality, Country)$ serves as a prior knowledge for filtering the predicted tail entities belonging to the concept $Country$. More generally, the set of tail concepts in common sense triples associated with a given query triple $(h, r, ?)$ is defined as:
\begin{equation}
    \mathcal{C}_{t}\ =\{c_{ti} \vert (c_{hi},r,c_{ti}) \in \mathcal{C}_1\}
\end{equation}
where $c_{ti}$ denotes the tail concept in the $i$-th individual-form common sense triple $(c_{hi},r,c_{ti})$ containing the relation $r$ and the concept $c_{hi}$ corresponding to the head entity $h$.

Then, all entities falling within the concept set $\mathcal{C}_{t}$ are identified as entity candidates with higher confidence compared to other entities. In the fine-grained entity prediction phase, we generate triple candidates $\{(h,r,e_j)\}$ derived from each candidate entity $e_{j}$. Afterwards, we employ the score function $E_{ec}(h,r,e_j)$ consistent with that utilized during training to compute the score for each triple candidate $(h,r,e_j)$. 

Furthermore, all triple candidates are organized in descending order based on their computed scores. The top-ranked triple candidates are extracted as the final inference results. The automatically generated common sense significantly contributes to enhancing the accuracy of KGC tasks. The training and coarse-to-fine inference procedures of ECSE model are exhibited in Algorithm~\ref{alg3}.

However, in practical, it is acknowledged that not all KGs possess abundant entity concepts. Consequently, it is imperative to explore techniques specific to KGs lacking explicit entity concepts, with a view to enhancing the scalability of the common sense-enhanced KGC framework.

\section{Implicit Common Sense-Guided Model}
\label{sec:Auto}

With regard to the KGs that lack sufficient entity concepts, we extend the factual triples to be implicit common sense triples to guarantee the scalability of our framework. Consequently, we develop a relation-aware concept embedding module to learn embeddings of concepts and common sense-level relations. In order to obtain inference results, we score each candidate triple using a joint common sense and fact-driven fashion.

\begin{figure}
    \centering
    \includegraphics[scale=0.32]{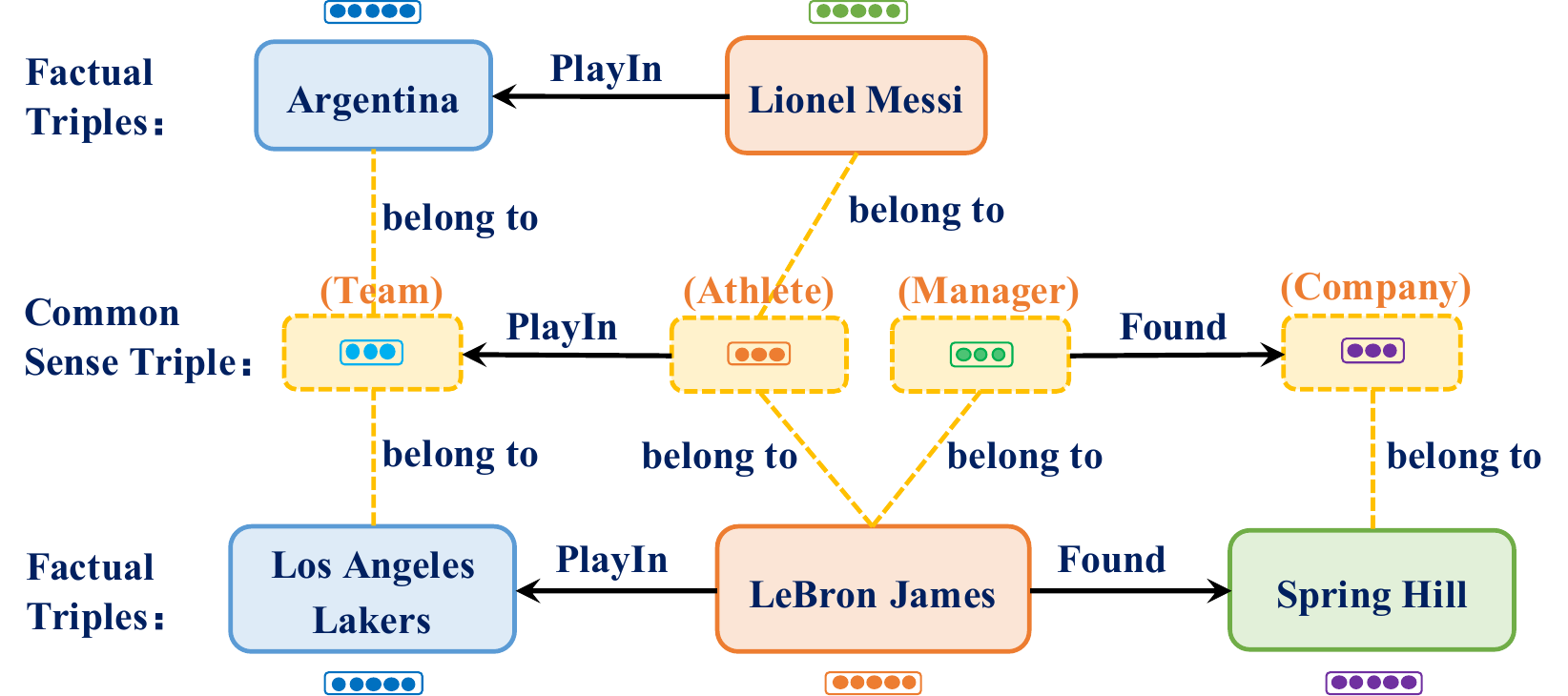}
    \caption{An illustration of factual triples along with their corresponding implicit common-sense triples. A factual triple comprises an explicit entity pair connected by a relation while a common sense triple involves implicit entity concepts along with an explicit relation. It is worth noting that The entity concepts in parentheses are intended to convey potential meanings rather than real textual representations.} 
    \label{fig:conceptembedding}
\end{figure}

\subsection{Relation-Aware Concept Embedding}


Each factual triple $(h, r, t)$ can be expanded into an implicit common sense triple $(c_h, r, c_t)$, where the head concept $c_h$ and the tail concept $c_t$ should imply the abstract representations of entities adaptive to the specific relation $r$. For instance, in Fig.~\ref{fig:conceptembedding}, the observed factual triples reveal that the entity $LeBron\ James$ is associated with the ontological concepts of $Athlete$ and $Manager$ through intuitive inference. These ontological associations are specific to the relations $PlayIn$ and $Found$, respectively. However, the entities $LeBron\ James$ and $Lionel\ Messi$ exhibit similar conceptual representations referred to the same relation $PlayIn$. Motivated by these observations, we propose a relation-aware concept embedding mechanism for modeling implicit common sense.

Primarily, we introduce the meta-concept embedding to represent each entity concept independent of any relation. Then, to accurately represent the entity concept that complies with the semantics of a specific relation, we design a relation-aware projection operator to transform the meta-concept embedding of an entity into the appropriate relation-aware concept embeddings adaptive to the associated relations. Then, we could model the implicit common sense triple with various common sense-specific score functions defined as follows:

(1) Translation-based score function:
\begin{align}
    \mathbf{c}_{h,r} = \mathbf{M}_r\ \mathbf{c}_h,\ \ \ \mathbf{c}_{t,r} = \mathbf{M}_r\ \mathbf{c}_t \label{eq11} \\
     E_{cs}(h,r,t) = -\Vert \mathbf{c}_{h,r} + \mathbf{c}_{r} - \mathbf{c}_{t,r} \Vert \label{eq12}
\end{align}
in which $\mathbf{c}_r$ denotes the vector embedding of the relation $r$ in the real space, $\mathbf{c}_{h}$ and $\mathbf{c}_{t}$ indicate the meta-concept embeddings of the head entity $h$ and the tail entity $t$, respectively. $\mathbf{c}_{h,r}$ and $\mathbf{c}_{t,r}$ are the concept embeddings of entities $h$ and $t$ adaptive to the relation $r$ via the projection matrix $\mathbf{M}_r$.

(2) Rotation-based score function:
\begin{align}
    \mathbf{c}_{h,r} = \mathbf{p}_r \circ \mathbf{c}_h,\ \ \ \mathbf{c}_{t,r} = \mathbf{p}_r \circ \mathbf{c}_t \label{eq14} \\
    E_{cs}(h,r,t) = -\Vert \mathbf{c}_{h,r} \circ \mathbf{c}_{r} - \mathbf{c}_{t,r} \Vert \label{eq15}
\end{align}
where $\mathbf{c}_r$, $\mathbf{c}_{h}$ and $\mathbf{c}_{t}$ denotes the similar embeddings in Eq.~\ref{eq11}-\ref{eq12} but are represented as the complex vectors. $\mathbf{c}_{h,r}$ and $\mathbf{c}_{t,r}$ are the concept embeddings of entities via the projection operator $\mathbf{p}_r$ in the complex vector space specific to the relation $r$.

(3) Tensor decomposition-based score function:
\begin{align}
    \mathbf{c}_{h,r} =\mathbf{p}_r \circ \mathbf{c}_h,\ \ \ \mathbf{c}_{t,r} = \mathbf{p}_r \circ \mathbf{c}_t \label{eq16} \\
    E_{cs}(h,r,t)={\rm Re}(\mathbf{c}_{h,r}^\top {\rm diag}\left(\mathbf{c}_r\right)\overline{\mathbf{c}}_{t,r}) \label{eq17}
\end{align}
where ${\rm diag}(\mathbf{c}_r)$ represents the diagonal matrix embedding in the complex space of the relation $r$. $\mathbf{p}_r$, $\mathbf{c}_h$, $\mathbf{c}_t$, $\mathbf{c}_{h,r}$ and $\mathbf{c}_{t,r}$ are the same definitions in Eq.~\ref{eq14} and Eq.~\ref{eq15}. $\overline{\mathbf{c}}_{t,r}$ indicates the conjugation of $\mathbf{c}_{t,r}$.

Actually, a common sense triple reveals the abstract and generalized semantics of the intersection among a relation and its connected entity pair. For common sense triples, the head entities associated with the relation $Found$ all refer to the concept $Manager$, whereas the tail entities are tied to the concept $Company$. Inspired by this natural property, we propose a concept similarity constraint mechanism to enhance the abstract features of concept embeddings. For two factual triples $\left(h_1,r,t_1\right)$ and $\left(h_2,r,t_2\right)$ with the same relation $r$, we expect that the concept embeddings in these two triples satisfy:
\begin{equation}
    \mathbf{c}_{h1,r}=\mathbf{c}_{h_2,r},\ \ \ \mathbf{c}_{t1,r}=\mathbf{c}_{t_2,r} \label{eq18}
\end{equation}
where $\mathbf{c}_{h1,r}$ and $\mathbf{c}_{h_2,r}$ denote two head concepts while $\mathbf{c}_{t1,r}$ and $\mathbf{c}_{t_2,r}$ represent two tail concepts associated with the relation $r$ obtained by relation-aware concept embedding.

Furthermore, regarding arbitrary factual triples $\left(h_1,r_1,t_1\right)$ and $\left(h_2,r_2,t_2\right)$, we define a score function for evaluating the similarity of concepts in these two triples as:
\begin{align}
    E_{sim}((h_1, r_1, t_1), (h_2, r_2, t_2)) = & - 0.5 \cdot (\Vert \mathbf{c}_{h1,r1} - \mathbf{c}_{h2,r2} \Vert \nonumber \\
    & + \Vert \mathbf{c}_{t1,r1} - \mathbf{c}_{t2,r2} \Vert)
\end{align}
in which $\mathbf{c}_{h1,r1}$ and $\mathbf{c}_{t1,r1}$ are the head and tail concept embeddings relevant to the relation $r_1$. $\mathbf{c}_{h2,r2}$ and $\mathbf{c}_{t2,r2}$ denote the head and tail concept embeddings adaptive to the relation $r_2$, respectively. In this regard, according to the expectation defined in Eq.~\ref{eq18}, the score function $E_{sim}$ would tend to a larger value when $r_1$ and $r_2$ are actually the same relations, which enhances the abstract feature of the entities to represent the semantics of concepts in implicit common sense.

\subsection{Joint Embedding Based on Common Sense and Fact}

The model-agnostic nature of our framework signifies that we could directly leverage the score functions of many existing KGE models to represent a factual triple $(h, r, t)$. To maintain consistency with the score functions in Eq.~\ref{eq11}-\ref{eq17}, the fact-specific score functions are clarified as:

(1) Translation-based score function (TransE):
\begin{equation}
     E_{f}(h,r,t) = -\Vert \mathbf{h} + \mathbf{r} - \mathbf{t} \Vert
\end{equation}
in which $\mathbf{h}$, $\mathbf{t}$ and $\mathbf{r}$ are vector embeddings in the real space of the entities $h$ and $t$ as well as the relation $r$.

(2) Rotation-based score function (modified RotatE):
\begin{align}
    \mathbf{h}_r = & \mathbf{h} - {\mathbf{w}_r}^\top \mathbf{h} \mathbf{w}_r, \ \ \ \mathbf{t}_r = \mathbf{t} - {\mathbf{w}_r}^\top \mathbf{t} \mathbf{w}_r \label{eq21}\\
    & E_{f}(h,r,t) = -\Vert \mathbf{h}_{r} \circ \mathbf{r} - \mathbf{t}_{r} \Vert \label{eq22}
\end{align}
where $\mathbf{h}$, $\mathbf{t}$ and $\mathbf{r}$ are complex vector embeddings of $h$, $t$ and $r$. $\mathbf{c}_{h,r}$ and $\mathbf{c}_{t,r}$ are the mapped entity embeddings obtained by a hyperplane with the normal vector $\mathbf{w}_r$ associated with the relation $r$.

(3) Tensor decomposition-based score function (ComplEx):
\begin{equation}
E_{f}(h,r,t)={\rm Re}(\mathbf{h}^\top {\rm diag}\left(\mathbf{r}\right)\overline{\mathbf{t}}) \label{eq23}
\end{equation}
where ${\rm diag}(\mathbf{r})$ is the diagonal matrix in the complex space for representing the relation $r$. $\mathbf{h}$ and $\mathbf{t}$ are the complex vector embeddings of $h$ and $t$. $\overline{\mathbf{t}}$ denotes the conjugation of $\mathbf{t}$.

Based on the previously defined score functions, we propose a joint learning approach tailored to both common sense triples and factual triples. The objective is to ensure that the learned entity embeddings capture unique characteristics, while concept embeddings emphasize abstract semantics. This is achieved by embedding entities in a higher-dimensional representation space compared to concepts. The fact-level embeddings of entities and relations, along with the common sense-level embeddings of concepts and relations, are achieved through a multi-task learning scheme. Besides, we employ a contrastive learning strategy to enhance the semantic similarity of concept embeddings within the same category. Consequently, the overall loss function is formulated as follows:
\begin{equation}
    L=\sum_{\left(h,r,t\right)\in T}\left\{\sum_{\left(h^\prime,r,t^\prime\right)\in T^\prime}\left\{L_{f}+\alpha_1 L_{cs}\right\}+\alpha_2 L_{sim}\right\} \label{eq24}
\end{equation}
in which the overall loss function $L$ consists of three components, namely fact-specific loss $L_f$, common sense-specific loss $L_{cs}$ and concept embedding similarity regularization $L_{sim}$. The weights $\alpha_1$ and $\alpha_2$ serve as trade-off parameters to balance the three components. $T$ denotes the set of all positive triples in the training set, while $T^\prime$ is the set of negative triples generated using self-adversarial negative sampling~\cite{RotatE}. The loss functions $L_f$, $L_{cs}$, and $L_{sim}$, are defined as follows:
\begin{align}
    &L_f=-\log{\sigma}\left(\gamma_1+E_{f}\left(h,r,t\right)\right) 
     -\log{\sigma}\left(-E_{f}\left(h^\prime,r,t^\prime\right)-\gamma_1\right) \label{eq25} \\
    &L_{cs}=max{\left[0, \gamma_2-E_{cs}\left(h,r,t\right)+E_{cs}\left(h^\prime,r,t^\prime\right)\right]} \label{eq26} \\
    &L_{sim}\ = \sum_{(h_p,r,t_p)\in Y}\sum_{(h_n,r^\prime,t_n)\in Y^\prime} max[0, \gamma_3-E_{sim}((h, r, t), \nonumber \\
    &\ \ \ \ \ \ \ \ \ \ \ \ (hp, r, tp))+E_{sim}((h, r, t), (hn, r^\prime, tn))] \label{eq27}
\end{align}
in which $\gamma_1$, $\gamma_2$ and $\gamma_3$ denote the margins in $L_f$, $L_{cs}$ and $L_{sim}$, respectively. $\sigma$ represents the sigmoid function. In Eq.~\ref{eq27}, the positive triple $\left(h,r,t\right)$ is regarded as an anchor instance. Meanwhile, $(h_p,r,t_p)$ denotes a positive instance within the set of all triples sharing the same relation $r$ with the anchor instance. $(h_n,r^\prime,t_n)$ signifies a negative instance in the set $Y^\prime$ containing the triples without the relation $r$.

The embeddings of fact-level entities and relations together with common sense-level concepts and relations can be learned by optimizing the loss functions in Eq.~\ref{eq24}-Eq.~\ref{eq27}. During the inference stage, the plausibility of each triple candidate $(h, r, e_j)$ is assessed by a dual score function from both the fact and common sense perspectives, which is formulated as:
\begin{equation}
    E_{ic}\left(h,r,e_j\right)=E_{f}\left(h,r,e_j\right)+\alpha_1 E_{cs}\left(h,r,e_j\right) \label{eq28}
\end{equation}
 
The intuition behind this scoring function is that a triple is more likely to be true if it is consistent with both the common sense and the factual knowledge. Then, all triple candidates are sorted in descending order based on their scores derived from Eq.~\ref{eq28}, and the candidate triples with higher rankings are considered as the inference results. Importantly, ICSE approach can serve as a pluggable module to be conveniently integrated with any KGE model on any KG to exploit implicit common sense for improving KGC performance.


\subsection{Proof of Representing Various Relational Characteristics}


Naturally, there are four relational patterns in most KGs, namely symmetry, anti-symmetry, inversion, and composition. Additionally, relations exhibit complex properties including 1-1, 1-N, N-1, and N-N. For a clearer understanding, we offer definitions and examples of relational patterns and complex properties in Table~\ref{table1}. In this paper, we analyze the capability of our approach in effectively representing these relational patterns and complex relations. This capability plays a pivotal role in the overall inference performance of a KGE model. A comparison between some existing models and our approach in handling various relational patterns and complex properties of relations is presented in Table~\ref{table2}. Detailed proofs are provided in the following.

\begin{table*}
 \centering
 \renewcommand\tabcolsep{12.0pt}
 \caption{The definitions and cases of various relational patterns and complex properties.}
 \begin{tabular}{c|l|l}
 \toprule
Relational Characteristics		& Definition    & Case\\
 \midrule
 symmetry	& $(t,\ r,h)\Leftarrow\left(h,r,t\right)$  & If $(David, SpouseOf, Mary)$, then $(Mary, SpouseOf, David)$ \\
 \midrule
 anti-symmetry	& $\lnot (t,\ r,h)\Leftarrow (h,r,t)$  & If $(London, Capital, England)$, then $\lnot(England, Capital, London)$ \\
 \midrule
 inversion      & $\left(t,\ r_2,h\right)\Leftarrow\left(h,r_1,t\right)$    & If $(David, StudentOf, Bill)$, then $(Bill, TeacherOf, David)$ \\
 \midrule
 \multirow{2}{45pt}{composition}      & \multirow{2}{100pt}{$\left(e_1,r_3,e_3\right)\Leftarrow\left(e_1,r_1,e_2\right)\ \land\left(e_2,\ r_2,e_3\right)$}    & If $(Herry, SonOf, Bill)$ and $(Bill, SpouseOf, Anna)$, \\
 &  & then $(Herry, SonOf, Anna)$ \\
 \midrule
 \midrule
 1-N    & $\left(h,r,t_1\right)\ \land \left(h, r,t_2\right)$   & $(Trump, President, U.S.A.) \land (Biden, President, U.S.A.)$ \\
 \midrule
 N-1    & $\left(h_1,r,t\right) \land \left(h_2, r,t\right)$   & $(Herry, BirthPlace, London) \land (Anna, BirthPlace, London)$ \\
 \midrule
 \multirow{2}{18pt}{\ N-N}    & $\left(h_1,r,t_1\right) \land \left(h_1, r, t_2\right) \land$    & $(Messi, PlayIn, PSG F.C.) \land (Messi, PlayIn, Argentina) \land$\\
        & $\left(h_2, r, t_1\right) \land \left(h_3, r, t_1\right) \land \cdots$    & $(Fazio, PlayIn, Argentina) \land (Higuain, PlayIn, Argentina)$ \\
\bottomrule
 \end{tabular}
 \label{table1}
 \end{table*}

\begin{table}
 \centering
 \renewcommand\tabcolsep{7.0pt}
 \caption{Comparison on representing various relational characteristics. Sym: symmetry, Anti-smy: anti-symmetry, Inv: inversion, Compo: composition, ComP: complex propertities.}
 \begin{tabular}{c|c|c|c|c|c}
 \toprule
 Model		& Sym    & Anti-sym   & Inv     & Compo   & ComP \\
 \midrule
 TransE~\cite{Bordes:TransE}	    & $\times$      & $\checkmark$      & $\checkmark$    & $\checkmark$    & $\times$ \\
 TransH~\cite{Wang:TransH}	    & $\times$      & $\checkmark$      & $\checkmark $   & $\checkmark$    & $\checkmark$ \\
 DistMult~\cite{Distmult}	& $\checkmark $ & $\times$          & $\times$        & $\times$        & $\times$ \\
 ComplEx~\cite{Trouillon:ComplEx}	    & $\checkmark$  & $\checkmark$      & $\checkmark$    & $\times$        & $\checkmark$ \\
 RotatE~\cite{RotatE}	    & $\checkmark$  & $\checkmark$      & $\checkmark$    & $\checkmark$    & $\times$ \\
 Ours        & $\checkmark$  & $\checkmark$      & $\checkmark$    & $\checkmark$    & $\checkmark$ \\
\bottomrule
 \end{tabular}
 \label{table2}
 \end{table}

Lemma 1. Our model could represent the symmetry of relations specific to factual triples.
\begin{proof}
    Based on the score function defined in Eq~\ref{eq22}, two factual triples $(h,r,t)$ and $(t,r,h)$ associated with a symmetric relation $r$ should satisfy the following constraints:
    \begin{align}
        \mathbf{h}_r \circ \mathbf{r} = \mathbf{t}_r,\ \ \ \mathbf{t}_r \circ \mathbf{r} = \mathbf{h}_r \label{eq30}
    \end{align}
    From Eq.~\ref{eq30}, we can retrive that
    \begin{equation}
        \mathbf{h}_r \circ \mathbf{r} \circ \mathbf{r} = \mathbf{h}_r \label{eq31}
    \end{equation}
    Based on Eq.~\ref{eq31}, we deduce that $\mathbf{r}\circ\mathbf{r}=\mathbf{v}(1)$, where the notation $\mathbf{v}(1)$ signifies a vector whose elements are all 1. Thus, we could represent the symmetric relation $r$ for factual triples following this constraint of relation embedding.
\end{proof}

Lemma 2. Our model could represent the anti-symmetry of relations specific to factual triples.
\begin{proof}
    Given a symmetric relation $r$, two factual triples $\left(h,r,t\right)$ and $\lnot \left(t,r,h\right)$ hold. These two factual triples should satisfy the following constraints by score function in Eq.~\ref{eq22}:
    \begin{align}
        \mathbf{h}_r \circ \mathbf{r} = \mathbf{t}_r,\ \ \ \mathbf{t}_r \circ \mathbf{r} \neq \mathbf{h}_r \label{eq34}
    \end{align}
    According to Eq.~\ref{eq34}, we can deduce that
    \begin{equation}
        \mathbf{h}_r \circ \mathbf{r} \circ \mathbf{r} \neq \mathbf{h}_r \label{eq35}
    \end{equation}
    We could further obtain that $\mathbf{r} \circ \mathbf{r} \neq \mathbf{v}(1)$. Therefore, the embedding of any symmetric relation should satisfy the constraint $\mathbf{r}\circ\mathbf{r} \neq \mathbf{v}(1)$ for factual triples.
\end{proof}

Lemma 3. Our model could represent inverse relations specific to factual triples.
\begin{proof}
    If a relation $r_1$ is the inversion of a relation $r_2$, two factual triples $(h,r_1,t)$ and $(t,r_2,h)$ should satisfy the following constraints based on Eq~\ref{eq22}:
    \begin{align}
        \mathbf{h}_{r1} \circ \mathbf{r}_1 = \mathbf{t}_{r1},\ \ \ \mathbf{t}_{r2} \circ \mathbf{r}_2 = \mathbf{h}_{r2} \label{eq37}
    \end{align}
    If the normal vectors defined in Eq.~\ref{eq21} associated with $r_1$ and $r_2$ satisfy $\mathbf{w}_{r1}=\mathbf{w}_{r2}$, we can deduce that $\mathbf{h}_{r1} = \mathbf{h}_{r2}$ and $\mathbf{t}_{r1}=\mathbf{t}_{r2}$. Therefore, according to Eq.~\ref{eq37}, we can obtain that
    \begin{equation}
        \mathbf{h}_{r1} \circ \mathbf{r}_1 \circ \mathbf{r}_2 = \mathbf{h}_{r1} \label{eq38}
    \end{equation}
    From Eq.~\ref{eq38}, we could further deduce that $\mathbf{r}_1 \circ \mathbf{r}_2 = \mathbf{v}(1)$. Therefore, for factual triples, a relation $r_1$ and its inverse version $r_2$ should satisfy $\mathbf{r}_1 \circ \mathbf{r}_2 = \mathbf{v}(1)$ and $\mathbf{w}_{r1}=\mathbf{w}_{r2}$.
\end{proof}

Lemma 4. Our model could represent composite relations specific to factual triples.
\begin{proof}
    Three factual triples satisfy the logical association $\left(e_1,r_3,e_3\right)\Leftarrow\left(e_1,r_1,e_2\right)\ \land\left(e_2,\ r_2,e_3\right)$ if the relations $r_1$, $r_2$ and $r_3$ form the compositional pattern. The score functions of these triples following Eq.~\ref{eq22} are privided as:
    \begin{align}
        \mathbf{e}_{1,r1} \circ \mathbf{r}_1 = \mathbf{e}_{2,r1}, \mathbf{e}_{2,r2} \circ \mathbf{r}_2 = \mathbf{e}_{3,r2}, \mathbf{e}_{1,r3} \circ \mathbf{r}_3 = \mathbf{e}_{3,r3} \label{eq41}
    \end{align}
    If the normal vectors corresponding to the relations $r_1$, $r_2$ and $r_3$ satisfy $\mathbf{w}_{r1}=\mathbf{w}_{r2}=\mathbf{w}_{r3}$, the three equations in Eq.~\ref{eq41} can be rewritten as a single equation:
    \begin{equation}
        \mathbf{e}_{1,r1} \circ \mathbf{r}_1 \circ \mathbf{r}_2 = \mathbf{e}_{1,r1} \circ \mathbf{r}_3 \label{eq46}
    \end{equation}
    From Eq.~\ref{eq46}, we can obtain that $\mathbf{r}_1 \circ \mathbf{r}_2 = \mathbf{r}_3$. Thus, the relations $r_1$ and $r_2$ with their composite version $r_3$ in factual triples should satisfy $\mathbf{r}_1 \circ \mathbf{r}_2 = \mathbf{r}_3$ and $\mathbf{w}_{r1}=\mathbf{w}_{r2}=\mathbf{w}_{r3}$.
\end{proof}

Lemma 5. Our model could represent the complex properties of relations specific to factual triples.
\begin{proof}
    There are at least two factual triples $\left(h,r,t_1\right)$ and $\left(h,r,t_2\right)$ hold if $r$ is a 1-N relation. Based on the score function in Eq.~\ref{eq22}, these two triples should satisfy:
    \begin{align}
        \mathbf{h}_r \circ \mathbf{r} = \mathbf{t}_{1,r},\ \ \ \mathbf{h}_r \circ \mathbf{r} = \mathbf{t}_{2,r} \label{eq48}
    \end{align}
    From Eq.~\ref{eq48}, we acquire that $\mathbf{t}_{1,r} = \mathbf{t}_{2,r}$. Then, following the hyperplane defined in Eq.~\ref{eq21}, we could retrieve that
    \begin{equation}
        \mathbf{t}_1 - \mathbf{t}_1^\top \mathbf{w}_r \mathbf{t}_1=\mathbf{t}_2 - \mathbf{t}_2^\top \mathbf{w}_r \mathbf{t}_2 \label{eq49}
    \end{equation}
    From Eq.~\ref{eq49}, we can directly deduce the following constraint:
    \begin{equation}
        \mathbf{t}_1 - \mathbf{t}_2= \mathbf{t}_1^\top \mathbf{w}_r \mathbf{t}_1 - \mathbf{t}_2^\top \mathbf{w}_r \mathbf{t}_2 \label{eq50}
    \end{equation}
    Thus, the hyperplane parameter of 1-N relation $\mathbf{w}_r$ and the associated two tail entities conform to the constraint shown in Eq.~\ref{eq50}. Modeling complex properties of N-1 and N-N relations both follow the same analysis procedure.
\end{proof}

Lemma 6. Our model could represent the symmetric relations specific to common sense triples.
\begin{proof}
    Based on the score function specific to common sense triples defined in Eq.~\ref{eq12}, the common sense-level triples corresponding to two factual triples $\left(h,r,t\right)$ and $\left(t,r,h\right)$ containing a symmetric $r$ should satisfy the following constraints:
    \begin{align}
        \mathbf{c}_{h,r} + \mathbf{c}_r = \mathbf{c}_{t,r},\ \ \ \mathbf{c}_{t,r} + \mathbf{c}_r = \mathbf{c}_{h,r} \label{eq52}
    \end{align}
    On account of the two equations in Eq.~\ref{eq52}, we can obtain that
    \begin{equation}
        \mathbf{c}_{h,r} + \mathbf{c}_r + \mathbf{c}_r = \mathbf{c}_{h,r} \label{eq53}
    \end{equation}
    From Eq.~\ref{eq53}, we could further retrieve that
    \begin{equation}
        \mathbf{c}_r = \mathbf{v}(0),\ \ \ \mathbf{c}_{h,r}=\mathbf{c}_{t,r} \label{eq54}
    \end{equation}
    where the notation $\mathbf{v}(0)$ indicates a zero vector with the same dimension as $\mathbf{c}_r$. The finding in Eq.~\ref{eq54} is reasonable since the concepts of an entity pair linked by a symmetric relation logically are the same. For instance, both the head and tail entities associated with the symmetric relation $SpouseOf$ naturally belong to the concept $Person$ as shown in Table~\ref{table1}.
\end{proof}

Lemma 7. Our model could represent the anti-symmetric relations specific to common sense triples.
\begin{proof}
    Based on Eq.~\ref{eq12}, the score functions of common sense-level triples corresponding to two factual triples $\left(h,r,t\right)$ and $\lnot \left(t,r,h\right)$ containing a anti-symmetric $r$ are given as:
    \begin{align}
        \mathbf{c}_{h,r} + \mathbf{c}_r = \mathbf{c}_{t,r},\ \ \ \mathbf{c}_{t,r} + \mathbf{c}_r \neq \mathbf{c}_{h,r} \label{eq56}
    \end{align}
    From Eq.~\ref{eq56}, we could deduce that
    \begin{equation}
        \mathbf{c}_{h,r} + \mathbf{c}_r + \mathbf{c}_r \neq \mathbf{c}_{h,r} \label{eq57}
    \end{equation}
    Based on Eq.~\ref{eq57}, we further obtain that $\mathbf{c}_r \neq \mathbf{v}(0)$. As can be observed, the value of $\Vert \mathbf{c}_r \Vert$ is larger, the anti-symmetric relation can be represented more effectively.
\end{proof}

Lemma 8. Our model is able to model the relations of inversion specific to common sense triples.
\begin{proof}
    Based on the score functions defined in Eq.~\ref{eq11} and Eq.~\ref{eq12}, the common sense triples corresponding to two factual triples $\left(h,r_1,t\right)$ and $\left(t,r_2,h\right)$ with regard to the relation $r_1$ and its inverse version $r_2$ can be represented as:
    \begin{align}
        \mathbf{M}_{r1} \mathbf{c}_{h} + \mathbf{c}_{r1} = \mathbf{M}_{r1} \mathbf{c}_{t},\ \ \ \mathbf{M}_{r2} \mathbf{c}_{t} + \mathbf{c}_{r2} = \mathbf{M}_{r2} \mathbf{c}_{h} \label{eq59}
    \end{align}
    Here, we can define a transformation matrix $\mathbf{P}$ satisfying:
    \begin{equation}
        \mathbf{M}_{r1} = \mathbf{P} \mathbf{M}_{r2} \label{eq60}
    \end{equation}
    Substituting Eq.~\ref{eq60} into Eq.~\ref{eq59}, we can obtain that
    \begin{equation}
        \mathbf{M}_{r1} \mathbf{c}_{t} + \mathbf{P} \mathbf{c}_{r2} + \mathbf{c}_{r1} = \mathbf{M}_{r1} \mathbf{c}_{t} \label{eq62}
    \end{equation}
    From Eq.~\ref{eq62}, we directly deduce the constraint $\mathbf{c}_{r1} = - \mathbf{P} \mathbf{c}_{r2}$, which should be satisfied by the embeddings of two inverse relations for common sense triples.
\end{proof}

Lemma 9. Our model could represent the relations of composition pattern specific to common sense triples.
\begin{proof}
    Based on the score functions in Eq.~\ref{eq11} and Eq.~\ref{eq12}, the common sense triples corresponding to three factual triples $(e_1,r_1,e_2)$, $(e_2,r_2,e_3)$ and $(e_1,r_3,e_3)$ with the composition pattern can be represented as:
    \begin{align}
        \mathbf{M}_{r1} \mathbf{c}_{e1} + \mathbf{c}_{r1} &= \mathbf{M}_{r1} \mathbf{c}_{e2} \label{eq63} \\
        \mathbf{M}_{r2} \mathbf{c}_{e2} + \mathbf{c}_{r2} &= \mathbf{M}_{r2} \mathbf{c}_{e3} \label{eq64} \\
        \mathbf{M}_{r3} \mathbf{c}_{e1} + \mathbf{c}_{r3} &= \mathbf{M}_{r3} \mathbf{c}_{e3} \label{eq65}
    \end{align}
    Here, two transformation matrices $\mathbf{P}$ and $\mathbf{Q}$ are defined as:
    \begin{align}
        \mathbf{M}_{r3} = \mathbf{P} \mathbf{M}_{r1}, \ \ \ \mathbf{M}_{r3} = \mathbf{Q} \mathbf{M}_{r2} \label{eq67}
    \end{align}
    Substituting the two equations defined in Eq.~\ref{eq67} into Eq.~\ref{eq63} and Eq.~\ref{eq64} respectively, we could achieve that
    \begin{align}
        \mathbf{M}_{r3} \mathbf{c}_{e1} + \mathbf{P} \mathbf{c}_{r1} = \mathbf{M}_{r3} \mathbf{c}_{e2},\ \ \ \mathbf{M}_{r3} \mathbf{c}_{e2} + \mathbf{Q} \mathbf{c}_{r2} = \mathbf{M}_{r3} \mathbf{c}_{e3} \label{eq68}
    \end{align}
    Furthermore, combining the three equations in Eq.~\ref{eq68} and Eq.~\ref{eq65}, we can deduce that
    \begin{equation}
        \mathbf{M}_{r3} \mathbf{c}_{e1} + \mathbf{P} \mathbf{c}_{r1} + \mathbf{Q} \mathbf{c}_{r2} = \mathbf{M}_{r3} \mathbf{c}_{e1} + \mathbf{c}_{r3} \label{eq71}
    \end{equation}
    From Eq.~\ref{eq71}, it is retrieved that 
    \begin{equation}
        \mathbf{c}_{r3} = - \mathbf{P} \mathbf{c}_{r1} + \mathbf{Q} \mathbf{c}_{r2} \label{eq72}
    \end{equation}
    Therefore, the relations that form the composition pattern should satisfy the constraint $\mathbf{c}_{r3} = - \mathbf{P} \mathbf{c}_{r1} + \mathbf{Q} \mathbf{c}_{r2}$.
\end{proof}

\begin{table}
 \centering
 \renewcommand\tabcolsep{3pt}
 \caption{Statistics of the experimental datasets.}
 \begin{tabular}{c|ccc|ccc}
 \toprule
Dataset		& \#Relation	& \#Entity	& \#Concept	 & \#Train	& \#Valid	& \#Test \\
 \midrule
 FB15K	   & 1,345		   & 14,951    & 89        & 483,142	& 50,000	& 59,071 \\
 FB15K237   & 237           & 14,505    & 89        & 272,115   & 17,535    & 20,466 \\
 NELL-995   & 200           & 75,492    & 270       & 123,370   & 15,000    & 15,838 \\
 DBpedia-242 & 298          & 99,744    & 242       & 592,654   & 35,851    & 30,000 \\
 \midrule
 WN18       &18	           &40,943	   &-	       & 141,442   & 5,000	   & 5,000\\
 YAGO3-10 &37	&123,182	&-	&1,079,040	&5,000	&5,000 \\
 \bottomrule
 \end{tabular}
 \label{table3}
 \end{table}

Lemma 10. Our model could represent the complex properties of relations specific to common sense triples.
\begin{proof}
    The common sense triples corresponding to two factual triples $\left(h,r,t_1\right)$ and $\left(h,r,t_2\right)$ containing the 1-N relation $r$ are represented as followings by Eq.~\ref{eq11} and Eq.~\ref{eq12}:
    \begin{align}
        \mathbf{M}_{r} \mathbf{c}_{h} + \mathbf{c}_{r} = \mathbf{M}_{r} \mathbf{c}_{t1},\ \ \ \mathbf{M}_{r} \mathbf{c}_{h} + \mathbf{c}_{r} = \mathbf{M}_{r} \mathbf{c}_{t2} \label{eq74}
    \end{align}
    Combining the equations in Eq.~\ref{eq74}, we obtain that
    \begin{equation}
        \mathbf{M}_r\mathbf{y}_{t1} = \mathbf{M}_r\mathbf{y}_{t2} \label{eq75}
    \end{equation}
    Thus, a 1-N relation and the associated two tail entities follow the constraint in Eq.~\ref{eq75}. The analyses of representing N-1 and N-N relations specific to common sense triples are in the same way as that of 1-N relations.
\end{proof}

\section{Experiments}

In this section, the extensive expriments of KGC tasks are performed on several datasets. We analyze the performance of our framework in two modes namely ECSE and ICSE compared with various baseline models. Besides, we compare our common sense-guided negative sampling strategy with several previous negative sampling approaches. Furthermore, ablation studies and case studies are conducted to illustrate the effectiveness of our framework incorporating common sense into KGC tasks.

\begin{table*}
\centering
\caption{Comparison results of baselines and two versions of our framework. The best results are bold and the second best ones are underlined. ``GA'' indicates the performance gain achieved by our approach compared with the best-performing baseline model$^{\blacktriangle}$.}
\renewcommand\tabcolsep{7.0pt}
\begin{tabular}{c|ccccc|ccccc}
\toprule
\multirow{2}*{Models} & \multicolumn{5}{c|}{FB15K}   & \multicolumn{5}{c}{FB15K237}\\
& MR $\downarrow$     & MRR $\uparrow$      & Hits@10 $\uparrow$       & Hits@3 $\uparrow$       & Hits@1 $\uparrow$    & MR $\downarrow$     & MRR $\uparrow$      & Hits@10 $\uparrow$    & Hits@3 $\uparrow$    & Hits@1 $\uparrow$ \\
\midrule
TransE~\cite{Bordes:TransE}     & 35   & 0.626     & 0.838     & 0.723     & 0.496       & 195     & 0.268     & 0.454     & 0.298     & 0.176 \\
TransH~\cite{Wang:TransH}     & 90   & 0.496     & 0.754     & 0.616     & 0.334       & 239     & 0.283     & 0.482     & 0.325     & 0.183 \\
TransR~\cite{Lin:TransR}     & 88   & 0.532     & 0.774     & 0.650     & 0.378       & 276     & 0.308    & 0.503     & 0.347$^{\blacktriangle}$     & 0.211 \\
DistMult~\cite{Distmult}       & 46   & 0.499     & 0.734     & 0.576     & 0.369       & 230     & 0.307     & 0.507$^{\blacktriangle}$     & 0.342     & 0.209 \\
HolE~\cite{HolE}                & 43   		& 0.524     & 0.739     & 0.613     & 0.402   & 545      & 0.238    & 0.431  & 0.331  & 0.144 \\
SimplE~\cite{SimplE}     & 199   & 0.222     & 0.438     & 0.247     & 0.121       & 433     & 0.183     & 0.352     & 0.199     & 0.103 \\
ComplEx~\cite{Trouillon:ComplEx}  & 41   & 0.556     & 0.786     & 0.641     & 0.425       & 197     & 0.265     & 0.434     & 0.291     & 0.182 \\
RotatE~\cite{RotatE}     & 35   & 0.657     & 0.850     & 0.746     & 0.537      & 204     & 0.269     & 0.452     & 0.298     & 0.179 \\
PairRE~\cite{PairRE}  & 79     & 0.573     & 0.764     & 0.652     & 0.457   & 193   & 0.310$^{\blacktriangle}$     & 0.483     & 0.340     & 0.223$^{\blacktriangle}$ \\
HAKE~\cite{HAKE}   & \underline{34}$^{\blacktriangle}$    & 0.690$^{\blacktriangle}$     & 0.872$^{\blacktriangle}$     & 0.780$^{\blacktriangle}$     & 0.574$^{\blacktriangle}$      & 176$^{\blacktriangle}$     & 0.306     & 0.486     & 0.337     & 0.216 \\
\midrule
HAKE+\textbf{ECSE}	        & \textbf{30}   & \textbf{0.741}    & \textbf{0.896}    & \textbf{0.825}	& \textbf{0.646}     & \underline{170} & \underline{0.321}  & \underline{0.515}      & \underline{0.355}   & \underline{0.227}  \\
HAKE+\textbf{ICSE}	        & 36   & \underline{0.701}    & \underline{0.881}    & \underline{0.795}	& \underline{0.583}     & \textbf{169} & \textbf{0.330}  & \textbf{0.528}      & \textbf{0.365}   & \textbf{0.232}  \\
\midrule
GA	        & 13.3\%   & 7.4\%    & 2.8\%    & 5.8\%	& 12.5\%     & 4.1\% & 6.5\%  & 4.1\%      & 5.2\%   & 4.0\%  \\
\bottomrule
\toprule
\multirow{2}*{Models} & \multicolumn{5}{c|}{DBpedia-242}   & \multicolumn{5}{c}{NELL-995}\\
& MR $\downarrow$     & MRR $\uparrow$      & Hits@10 $\uparrow$       & Hits@3 $\uparrow$       & Hits@1 $\uparrow$    & MR $\downarrow$     & MRR $\uparrow$      & Hits@10 $\uparrow$    & Hits@3 $\uparrow$    & Hits@1 $\uparrow$ \\
\midrule
TransE~\cite{Bordes:TransE}     & 2733   & 0.242     & 0.468    & 0.344     & 0.100    & 1081      & 0.429     & 0.557     & 0.477      & 0.354 \\
TransH~\cite{Wang:TransH}     & \underline{1402}$^{\blacktriangle}$   & 0.298     & 0.564     & 0.422     & 0.130       & \underline{826}$^{\blacktriangle}$     & 0.446     & 0.565     & 0.489     & 0.372 \\
TransR~\cite{Lin:TransR}     & 2337   & 0.152     & 0.427     & 0.204     & 0.025       & 8802     & 0.097    & 0.233     & 0.129     & 0.021 \\
DistMult~\cite{Distmult}     & 12289      & 0.195     & 0.337     & 0.229     & 0.122       & 7241     & 0.165     & 0.250     & 0.172     & 0.122 \\
HolE~\cite{HolE}                & 11251   		& 0.168     & 0.321     & 0.258     & 0.086   & 14796      & 0.176    & 0.278  & 0.227  & 0.125 \\
SimplE~\cite{SimplE}     & 3899   & 0.115     & 0.251     & 0.122     & 0.050       & 12747     & 0.068     & 0.168     & 0.066     & 0.025 \\
ComplEx~\cite{Trouillon:ComplEx}  & 2750   & 0.141     & 0.241    & 0.155     & 0.088    & 6427     & 0.191     & 0.277     & 0.212      & 0.144 \\
RotatE~\cite{RotatE}     & 1950   & 0.374     & \underline{0.582}$^{\blacktriangle}$     & 0.457     & 0.249        & 2077       & 0.460     & 0.553    & 0.493  & 0.403 \\
PairRE~\cite{PairRE}  & 1593     & 0.333     & 0.566     & 0.423     & 0.199  & 1358   & 0.450     & 0.544     & 0.479     & 0.392 \\
HAKE~\cite{HAKE}   & 1665     & 0.408$^{\blacktriangle}$     & 0.579     & 0.463$^{\blacktriangle}$    & 0.312$^{\blacktriangle}$     & 1157    & 0.502$^{\blacktriangle}$     & 0.610$^{\blacktriangle}$      & 0.538$^{\blacktriangle}$   & 0.437$^{\blacktriangle}$ \\
\midrule
HAKE+\textbf{ECSE}	        & \textbf{931}   & \textbf{0.437}    & \textbf{0.593}    & \textbf{0.481}	& \textbf{0.353}     & \textbf{433} & \textbf{0.543}  & \textbf{0.655}      & \textbf{0.583}   & \textbf{0.477}  \\
HAKE+\textbf{ICSE}	        & 1782   & \underline{0.410}    & \underline{0.582}    & \underline{0.465}	& \underline{0.313}     & 943 & \underline{0.511}  & \underline{0.622}      & \underline{0.550}   & \underline{0.443}  \\
\midrule
GA	        & 50.6\%   & 7.1\%    & 1.9\%    & 3.9\%	& 13.1\%     & 90.8\% & 8.2\%  & 7.4\%      & 8.3\%   & 9.2\%  \\
\bottomrule
\end{tabular}
\label{table4}
\end{table*}

\subsection{Experimental Settings}

\subsubsection{\textbf{Datasets}}

Our experiments leverage six commonly-used benchmark datasets, which can be categorized into the following two groups. (1) Datasets with entity concepts: FB15K~\cite{Bordes:TransE} and FB15K237~\cite{FB15k237} are two subsets of Freebase~\cite{BGF:Freebase} where each entity always belongs to multiple concepts. NELL-995~\cite{DeepPath} is sampled from NELL~\cite{Mitchell:nell}. DBpedia-242~\cite{EngineKG} is extracted from DBpedia~\cite{Lehmann:dbpedia}. Particularly, each entity in NELL-995 and DBpedia-242 is linked by a single concept. (2) Datasets without entity concepts: WN18~\cite{Bordes:TransE} is a dataset extracted from NELL~\cite{Mitchell:nell}. YAGO3-10~\cite{Dettmers:ConvE} is the subset of YAGO~\cite{Yago}. The detailed statistics for these six datasets are provided in Table~\ref{table3}.


\subsubsection{\textbf{Baselines}}

We compare our framework with several typical and state-of-the-art KGE baseline approaches, including TransE~\cite{Bordes:TransE}, TransH~\cite{Wang:TransH}, TransR~\cite{Lin:TransR}, DistMult~\cite{Distmult}, HolE~\cite{HolE}, SimplE~\cite{SimplE},  ComplEx~\cite{Trouillon:ComplEx}, RotatE~\cite{RotatE}, PairRE~\cite{PairRE} and HAKE~\cite{HAKE}. We select these baselines rather than some most recent models since our framework mainly aims to enhance the performance of existing models. For a comparison, we reuse the released source codes$\footnote{TransE/DistMult/ComplEx/RotatE: \url{https://github.com/DeepGraphLearning/KnowledgeGraphEmbedding}, TransH/TransR/HolE/SimplE: \url{https://github.com/thunlp/OpenKE}, PairRE: \url{https://github.com/ant-research/KnowledgeGraphEmbeddingsViaPairedRelationVectors_PairRE}, 
HAKE: \url{https://github.com/MIRALab-USTC/KGE-HAKE}.}$ of these baseline models to achieve their evaluation results on KGC tasks.

We select these baseline models for several reasons: (1) These models have exhibited strong performance on KGC tasks in prior researches, making them suitable for meaningful comparisons in our experiments. (2) They are frequently-used as baselines for KGC. (3) Many of these baselines have publicly available source codes, facilitating their integration into our framework and ensuring a fair evaluation.

\subsubsection{\textbf{Evaluation Protocol}}

The experiments are implemented using PyTorch on a system running Ubuntu 16.04, equipped with an Intel i9-9900 CPU and a GeForce GTX 2080Ti GPU. Hyper-parameter tuning is performed on the validation set to optimize model performance. The following hyper-parameters are considered during tuning: dimensions of embeddings are selected from the set: $\{150, 500, 1000\}$. Learning rate is chosen from the range: $\{0.0001, 0.001, 0.005, 0.01\}$. Margins $\gamma$, $\gamma_1$, $\gamma_2$, and $\gamma_3$ are selected from: $\{9,12,18,24,30\}$. Sampling temperature in negative sampling is set to either 0.5 or 1.0. \textbf{It is noteworthy that the negative sampling size is fixed at 2 for all the baselines and our models.} The other hyper-parameters are selected based on the recommended settings in the respective original papers. For a fair comparison, we maintain consistent settings across all models in the experiments.

\begin{table*}[!t]
\centering
\renewcommand\tabcolsep{10pt}
\caption{Comparison results of our ECSE model and baselines on four datasets. ``+CGNS'', ``+CFI'' and ``+ECSE'' indicate the integration of the basic model with common sense-guided negative sampling, coarse-to-fine inference and the whole ECSE model, respectively.}
\begin{tabular}{l|ccccc|ccccc}
\toprule
\multirow{2}*{Models} & \multicolumn{5}{c|}{FB15K} & \multicolumn{5}{c}{FB15K237} \\
	& MR	& MRR	& Hits@10	& Hits@3  & Hits@1	& MR	& MRR	& Hits@10	& Hits@3  & Hits@1\\
\midrule
TransE~\cite{Bordes:TransE}         & 35   & 0.626     & 0.838     & 0.723     & 0.496       & 195     & 0.268     & 0.454     & 0.298     & 0.176 \\
TransE+\textbf{CGNS}	        & 34   & 0.671    & 0.864    & \textbf{0.761}	& 0.552     & \textbf{175} & 0.298  & 0.490      & 0.333   & 0.203  \\
TransE+\textbf{CFI}	        & 35   & 0.636    & 0.839    & 0.725	&0.513     & 181 & 0.290  & 0.476      & 0.323   & 0.186  \\
TransE+\textbf{ECSE}	        & \textbf{33}   & \textbf{0.672}    & \textbf{0.865}    & \textbf{0.761}	& \textbf{0.555}     & \textbf{175} & \textbf{0.301}  & \textbf{0.493}      & \textbf{0.335}   & \textbf{0.206}  \\
\midrule
RotatE~\cite{RotatE}    & 35   & 0.657     & 0.850     & 0.746     & 0.537      & 204     & 0.269     & 0.452     & 0.298     & 0.179 \\
RotatE+\textbf{CGNS}	        & 33   & 0.702    & 0.877       & 0.790    & 0.588     & 182   & 0.296    & 0.486    & 0.329	& 0.202   \\
RotatE+\textbf{CFI}	        & 34   & 0.688    & 0.860       & 0.768    & 0.579	    & 188 & 0.308  & 0.493      & 0.340   & 0.217  \\
RotatE+\textbf{ECSE}	        & \textbf{31}   & \textbf{0.705}    & \textbf{0.878}    & \textbf{0.792}	& \textbf{0.593}     & \textbf{181} & \textbf{0.318}  & \textbf{0.511}      & \textbf{0.354}   & \textbf{0.223}  \\
\midrule
HAKE~\cite{HAKE}                    & 34    & 0.690     & 0.872     & 0.780     & 0.574      & 176     & 0.306     & 0.486     & 0.337     & 0.216 \\
HAKE+\textbf{CGNS}	        & 37   & 0.723    & 0.882    & 0.808	& 0.616     & 174 & 0.315  & 0.501      & 0.344   & 0.221  \\
HAKE+\textbf{CFI}	        & 32   & 0.729    & 0.890    & 0.817	& 0.622     & 172 & 0.320  & 0.508      & 0.352   & 0.226  \\
HAKE+\textbf{ECSE}	        & \textbf{30}   & \textbf{0.741}    & \textbf{0.896}    & \textbf{0.825}	& \textbf{0.646}     & \textbf{170} & \textbf{0.321}  & \textbf{0.515}      & \textbf{0.355}   & \textbf{0.227}  \\
\bottomrule
\toprule
\multirow{2}*{Models} & \multicolumn{5}{c|}{DBpedia-242} & \multicolumn{5}{c}{NELL-995} \\
	& MR	& MRR	& Hits@10	& Hits@3  & Hits@1	& MR	& MRR	& Hits@10	& Hits@3  & Hits@1\\
\midrule
TransE~\cite{Bordes:TransE}         & 2733   & 0.242     & 0.468    & 0.344     & 0.100    & 1081      & 0.429     & 0.557     & 0.477      & 0.354 \\
TransE+\textbf{CGNS}	        & 1889   & 0.287    & 0.575    & 0.427	& 0.103      & 1022   & 0.433    & 0.591    & 0.495	& 0.336   \\
TransE+\textbf{CFI}	        & \textbf{881}   & 0.322    & 0.585    & 0.450	& 0.152      & 336   & 0.509    & 0.617    & 0.547	& 0.444  \\
TransE+\textbf{ECSE}	        & \textbf{881}   & \textbf{0.330}    & \textbf{0.595}    & \textbf{0.458}	& \textbf{0.160}  & \textbf{317}   & \textbf{0.533}    & \textbf{0.650}    & \textbf{0.578}	& \textbf{0.461} \\
\midrule
RotatE~\cite{RotatE}                & 1950   & 0.374     & 0.582     & 0.457     & 0.249        & 2077       & 0.460     & 0.553    & 0.493  & 0.403 \\
RotatE+\textbf{CGNS}	        & 1063   & 0.407    & 0.593    & 0.476	& 0.300      & 1097   & 0.531    & 0.644    & 0.573	& 0.461   \\
RotatE+\textbf{CFI}	       & \textbf{983}   & 0.393    & 0.594    & 0.474	& 0.273   & 356   & 0.519    & 0.628    & 0.564	& 0.447 \\
RotatE+\textbf{ECSE}	        & 1027    & \textbf{0.423}    & \textbf{0.603}    & \textbf{0.486}	& \textbf{0.320}      & \textbf{329}   & \textbf{0.546}    & \textbf{0.660}    & \textbf{0.592}	& \textbf{0.474}   \\
\midrule
HAKE~\cite{HAKE}                    & 1757     & 0.408     & 0.579     & 0.463    & 0.312     & 1157    & 0.502     & 0.610      & 0.538   & 0.437 \\
HAKE+\textbf{CGNS}	        & 1147   & 0.427    & 0.587    & 0.472	& 0.341     & 2011 & 0.520  & 0.640      & 0.556   & 0.451  \\
HAKE+\textbf{CFI}	        & 1083   & 0.411    & 0.580    & 0.463	& 0.319     & 478 & 0.510  & 0.614      & 0.551   & 0.444  \\
HAKE+\textbf{ECSE}	        & \textbf{931}   & \textbf{0.437}    & \textbf{0.593}    & \textbf{0.481}	& \textbf{0.353}     & \textbf{433} & \textbf{0.543}  & \textbf{0.655}      & \textbf{0.583}   & \textbf{0.477}  \\
\bottomrule
\end{tabular}
\label{table4-2}
\end{table*}

To assess the effectiveness of our model in handling complex properties of relations, we categorize each relation based on a criterion from a previous study~\cite{Wang:TransH}. For each relation $r$, we calculate two key metrics: (1) Average number of tail entities $ah_t$ for each head entity. (2) Average number of head entities $at_h$ for each tail entity. Based on these metrics, we classify each relation into one of the following categories:
\begin{itemize}
    \item $r$ is a 1-1 relation if $ah_t<1.5$ and $at_h < 1.5$.
    \item $r$ is a 1-N relation if $ah_t>1.5$ and $at_h<1.5$.
    \item $r$ is an N-1 relation if $ah_t<1.5$ and $at_h>1.5$.
    \item $r$ is an N-N relation if $ah_t > 1.5$ and $at_h > 1.5$.
\end{itemize}

Following the inference stage of our framework, we can obtain the rank of the correct triple for each test instance according to the score function $E_{ec}$ defined in Eq.~\ref{eq3}-Eq.~\ref{eq5} or the score function $E_{ic}$ defined in Eq.~\ref{eq28}. Then, the performance of link prediction is evaluated by three commonly-used metrics: 
Following the inference stage of our framework, the performance of KG completion is assessed using the following three commonly-used metrics:
\begin{itemize}
    \item MR: mean rank of all the correct triples, which is calculated by
    \begin{equation}
        MR=\frac{1}{N}\cdot\sum_{i}^{N}{rank}_i \label{76}
    \end{equation}
where ${rank}_i$ represents the rank of the correct triple corresponding to the $i$-th test instance, and $N$ is the total number of test instances.

    \item MRR: mean reciprocal rank of all the correct triples, which can be computed via
    \begin{equation}
        MRR=\frac{1}{N}\cdot\sum_{i}^{N}\frac{1}{{rank}_i}
    \end{equation}

    \item Hits@K: proportion of the correct triples ranked in the top K, which could be obtained by
    \begin{equation}
        Hits@K=\frac{1}{N}\cdot\sum_{i}^{N}{{\mathbb{I}(rank}_i\le K)}
    \end{equation}
where the value of ${\mathbb{I}(rank}_i\le K)$ is 1 if ${rank}_i\le K$ is true. The value of $K$ is usually 1, 3, or 10.
\end{itemize}
 Particularly, the lower MR, the higher MRR and Hits@K reveals the better performance of KG completion. To ensure the validity of the evaluation results, all these metrics are computed in the filtered setting, which involves removing the triple candidates observed in the training sets from consideration.

 \subsection{Experimental Results}

 \subsubsection{\textbf{Results of Typical Baselines and Our Approaches}}
In this section, we present the global results of our framework compared with many typical baselines on the four datasets containing entity concepts. From the detailed results summarized in Table~\ref{table4}, our analysis reveals several key findings:

\begin{itemize}
    \item Consistent outstanding performance: on these four datasets, our models ECSE and ICSE consistently and significantly outperform all baselines. Specifically, ECSE and ICSE achieve both the best and second-best performance, illustrating the effectiveness of our common sense-enhanced KGC framework.
    \item Enhanced accuracy with common sense: notably, the ECSE model achieves the most significant performance gains on Hits@1 among the Hits@1/3/10 metrics on datasets FB15K, DBpedia-242, and NELL-995. This result demonstrates the superiority of supplementing common sense-level triples for enhancing the accuracy of KG completion tasks.
    \item Explicit concepts vs. Implicit concepts: ECSE performs better than ICSE on most datasets. This suggests that utilizing explicit concepts can represent common sense more accurately and improve the training and inference effectiveness of KGE models compared to employing implicit concepts.
\end{itemize}

These findings emphasize the effectiveness of our common sense-enhanced KGC framework, showcasing its potential to improve the performance of KGC tasks, particularly when exploiting explicit common sense with explicit concepts.

\subsubsection{\textbf{Results of Our Framework with Explicit Entity Concepts}}

To evaluate the performance of our ECSE model more specifically, we select three typical and well-performed KGE baseline models TransE, RotatE and HAKE as the basic modules of ECSE, which can be extended via ensembling the common sense-guided negative sampling (+CGNS), coarse-to-fine inference strategy (+CFI), and the entire ECSE model (+ECSE).

From the results shown in Table~\ref{table4-2}, we can observe that the performance of each basic model is obviously improved by CGNS or CFI modules. Furthermore, the ECSE framework consistently and significantly outperforms all the baselines, facilitating more performance improvements compared to each separate module. Compared to the average performance of the three baseline models, our ECSE model achieves improvements in MRR of 7.2\%, 11.5\%, 16.2\% and 16.7\% on FB15K, FB15K237, DBpedia-242 and NELL-995. These results demonstrate the superiority of integrating explicit common sense with various basic KGE models. 

\begin{table*}[!t]
\centering
\caption{Comparison results of our common sense-guided negative sampling and various existing negative sampling strategies.}
\renewcommand\tabcolsep{7pt}
\begin{tabular}{l|ccccc|ccccc}
\toprule
\multirow{2}*{Models} & \multicolumn{5}{c|}{FB15K} & \multicolumn{5}{c}{FB15K237} \\
	& MR	& MRR	& Hits@10	& Hits@3  & Hits@1	& MR	& MRR	& Hits@10	& Hits@3  & Hits@1\\
\midrule
TransE+Unifo~\cite{Bordes:TransE}         & 178 & 0.301     & 0.505     & 0.339     & 0.201     & 361    & 0.171    & 0.323     & 0.182     & 0.097 \\
TransE+NoSamp~\cite{nonesampling}        & 144  & 0.350     & 0.578     & 0.415     & 0.227       & 343      & 0.261    & 0.446     & 0.297     & 0.168 \\
TransE+NSCach~\cite{zhang2019nscaching}      & 209   & 0.292     & 0.560     & 0.375     & 0.144       & 556      & 0.205    & 0.353     & 0.226     & 0.131 \\
TransE+DomSam~\cite{DomainSampling}      & 35    & 0.619     & 0.839     & 0.715     & 0.489        & 186    & 0.283     & 0.467     & 0.314     & 0.190 \\
TransE+SAdv~\cite{RotatE}         & 35  & 0.626     & 0.838     & 0.723     & 0.496       & 195      & 0.268    & 0.454     & 0.298     & 0.176 \\
\midrule
\textbf{TransE+CGNS (Ours)}        & \textbf{34}   & \textbf{0.671}    & \textbf{0.864}    & \textbf{0.761}	& \textbf{0.552}    & \textbf{175}      & \textbf{0.298}    & \textbf{0.490}     & \textbf{0.333}     & \textbf{0.203} \\
\bottomrule
\toprule
\multirow{2}*{Models} & \multicolumn{5}{c|}{DBpedia-242} & \multicolumn{5}{c}{NELL-995} \\
	& MR	& MRR	& Hits@10	& Hits@3  & Hits@1	& MR	& MRR	& Hits@10	& Hits@3  & Hits@1\\
\midrule
TransE+Unifo~\cite{Bordes:TransE}         & 5750   & 0.124     & 0.262     & 0.183    & 0.033       & 8650    & 0.167    & 0.354     & 0.219     & 0.068 \\
TransE+NoSamp~\cite{nonesampling}        & 2292    & 0.202     & 0.395     & 0.247     & 0.101       & 9172    & 0.176    & 0.297     & 0.210     & 0.106 \\
TransE+NSCach~\cite{zhang2019nscaching}      & 5465   & 0.156     & 0.340     & 0.212     & 0.050       & 13967    & 0.107    & 0.205     & 0.122     & 0.107 \\
TransE+DomSam~\cite{DomainSampling}      & 3415   & 0.203     & 0.510     & 0.346     & 0.009      & 1319       & 0.381    & 0.549     & 0.468     & 0.271 \\
TransE+SAdv~\cite{RotatE}         & 2733   & 0.242     & 0.468    & 0.344     & 0.100     & 1081      & 0.429     & 0.557     & 0.477      & \textbf{0.354} \\
\midrule
\textbf{TransE+CGNS (Ours)}        & \textbf{1889}   & \textbf{0.287}    & \textbf{0.575}    & \textbf{0.427}	& \textbf{0.103}       & \textbf{1022}   & \textbf{0.433}    & \textbf{0.591}    & \textbf{0.495}	& 0.336 \\
\bottomrule
\end{tabular}
\label{table5}
\end{table*}

\begin{table*}[!t]
\renewcommand{\arraystretch}{0.6}
\centering
\renewcommand\tabcolsep{9pt}
\caption{Comparison results of our ICSE model and baselines on four datasets independent of explicit entity concepts. ``+ICSE'' indicates the integration of the basic model with the common sense-enhanced framework with implicit common sense.}
\begin{tabular}{l|ccccc|ccccc}
\toprule
\multirow{2}*{Models} & \multicolumn{5}{c|}{FB15K} & \multicolumn{5}{c}{FB15K237} \\
	& MR	& MRR	& Hits@10	& Hits@3  & Hits@1	& MR	& MRR	& Hits@10	& Hits@3  & Hits@1\\
\midrule
TransE~\cite{Bordes:TransE}         & 35   & 0.626     & 0.838     & 0.723     & 0.496       & 195     & 0.268     & 0.454     & 0.298     & 0.176 \\
 TransE+\textbf{ICSE}	        & \textbf{32}   & \textbf{0.668}    & \textbf{0.860}    & \textbf{0.754}	& \textbf{0.550}     & \textbf{165} & \textbf{0.323}  & \textbf{0.522}      & \textbf{0.361}   & \textbf{0.224}  \\
\midrule
ComplEx~\cite{Trouillon:ComplEx}    & \textbf{41}   & 0.556     & 0.786     & 0.641     & 0.425       & \textbf{187}     & 0.286     & 0.457     & 0.313     & 0.203 \\
ComplEx+\textbf{ICSE}	        & 45   & \textbf{0.645}    & \textbf{0.829}    & \textbf{0.724}	& \textbf{0.533}     & 240 & \textbf{0.298}  & \textbf{0.476}      & \textbf{0.325}   & \textbf{0.210}  \\
\midrule
RotatE~\cite{RotatE}    & 35   & 0.657     & 0.850     & 0.746     & 0.537      & 204     & 0.269     & 0.452     & 0.298     & 0.179 \\
RotatE+\textbf{ICSE}	        & \textbf{32}   & \textbf{0.680}    & \textbf{0.863}    & \textbf{0.770}	& \textbf{0.562}     & \textbf{172} & \textbf{0.323}  & \textbf{0.515}      & \textbf{0.357}   & \textbf{0.228}  \\
\midrule
HAKE~\cite{HAKE}                    & 34    & 0.690     & 0.872     & 0.780     & 0.574      & 176     & 0.306     & 0.486     & 0.337     & 0.216 \\
HAKE+\textbf{ICSE}               & \textbf{32}    & \textbf{0.701}     & \textbf{0.881}     & \textbf{0.795}     & \textbf{0.583}      & \textbf{170}     & \textbf{0.317}     & \textbf{0.506}     & \textbf{0.351}     & \textbf{0.223} \\
\bottomrule
\toprule
\multirow{2}*{Models} & \multicolumn{5}{c|}{WN18} & \multicolumn{5}{c}{YAGO3-10} \\
	& MR	& MRR	& Hits@10	& Hits@3  & Hits@1	& MR	& MRR	& Hits@10	& Hits@3  & Hits@1\\
\midrule
TransE~\cite{Bordes:TransE}         & 194   & 0.640     & 0.954    & 0.908     & 0.372    & 861      & 0.387     & 0.598     & 0.447      & 0.277 \\
TransE+\textbf{ICSE}	        & \textbf{180}   & \textbf{0.688}    & \textbf{0.955}    & \textbf{0.916}	& \textbf{0.461}  & \textbf{816}   & \textbf{0.476}    & \textbf{0.652}    & \textbf{0.530}	& \textbf{0.382} \\
\midrule
ComplEx~\cite{Trouillon:ComplEx}    & \textbf{347}   & 0.694     & 0.906    & 0.812     & 0.557    & \textbf{1473}     & 0.207     & 0.355     & 0.226      & 0.132 \\
ComplEx+\textbf{ICSE}	        & 886   & \textbf{0.940}    & \textbf{0.947}    & \textbf{0.944}	& \textbf{0.935}     & 3949 & \textbf{0.389}  & \textbf{0.574}      & \textbf{0.433}   & \textbf{0.295}  \\
\midrule
RotatE~\cite{RotatE}                & \textbf{202}   & 0.945     & 0.959     & 0.952     & 0.935        & 1321       & 0.389     & 0.591    & 0.434  & 0.288 \\
RotatE+\textbf{ICSE}	        & 219    & \textbf{0.950}    & \textbf{0.960}    & \textbf{0.953}	& \textbf{0.942}      & \textbf{1297}   & \textbf{0.469}    & \textbf{0.650}    & \textbf{0.524}	& \textbf{0.373}   \\
\midrule
HAKE~\cite{HAKE}               & \textbf{142}    & 0.917     & 0.952     & 0.930     & 0.898       & -     & -     & -     & -     & - \\
HAKE+\textbf{ICSE}          & 322    & \textbf{0.950}     & \textbf{0.962}     & \textbf{0.951}     & \textbf{0.942}      & -     & -     & -     & -     & - \\
\bottomrule
\end{tabular}
\label{table6}
\end{table*}

\begin{table*}
\centering
\caption{Evaluation results on FB15K and FB15K237 specific to complex properties of relations.}
\renewcommand\tabcolsep{15.0pt}
\begin{tabular}{c|cccc|cccc}
\toprule
\multicolumn{9}{c}{FB15K} \\
\midrule
\multirow{2}*{Models}  & \multicolumn{4}{c|}{Head Entity Prediction (Hits@10)}	 & \multicolumn{4}{c}{Tail Entity Prediction (Hits@10)}\\
	 & 1-1	& 1-N	& N-1	& N-N	 & 1-1	& 1-N	& N-1	& N-N\\
\midrule
TransE~\cite{Bordes:TransE}	& 0.886	& 0.969	& 0.577	& 0.833	 & 0.872	& 0.692	& 0.960	& 0.867\\
DistMult~\cite{Distmult}	& 0.864	& 0.957	& 0.517	& 0.714	 & 0.868	& 0.540	& 0.952	& 0.747\\
ComplEx~\cite{Trouillon:ComplEx} & 0.907	& 0.963	& 0.597	& 0.766	 & 0.904	& 0.660	& 0.954	& 0.799\\
RotatE~\cite{RotatE}	  & 0.922	& 0.969	& 0.602	& 0.839	 & 0.910	 & 0.677	& \underline{0.963}	& 0.875\\
PairRE~\cite{PairRE}	& 0.834	& 0.949	& 0.475	& 0.759	 & 0.827	& 0.528	& 0.947	& 0.793 \\
HAKE~\cite{HAKE}	& 0.921	& 0.970	& 0.649	& 0.868	 & 0.926	& 0.716	& \bf{0.964}	& 0.901\\
\midrule
HAKE+ECSE	& \underline{0.924}	& \underline{0.972}	& \bf{0.661}	& \bf{0.893}	 & \bf{0.930}	& \bf{0.734}	& \bf{0.964}	& \bf{0.923}\\
HAKE+ICSE	& \bf{0.928}	& \bf{0.974}	& 0.659	& 0.880	 & \underline{0.927}	& \underline{0.726}	& \bf{0.964}	& \underline{0.912}\\
\bottomrule
\toprule
\multicolumn{9}{c}{FB15K237} \\
\midrule
\multirow{2}*{Models}  & \multicolumn{4}{c|}{Head Entity Prediction (Hits@10)}	 & \multicolumn{4}{c}{Tail Entity Prediction (Hits@10)}\\
	 & 1-1	& 1-N	& N-1	& N-N	 & 1-1	& 1-N	& N-1	& N-N\\
\midrule
TransE~\cite{Bordes:TransE}	& 0.568	& 0.650	& 0.140	& 0.405	 & 0.563	& 0.117	& 0.871	& 0.552\\
DistMult~\cite{Distmult}	& 0.240	& 0.602	& 0.095	& 0.349	 & 0.229	& 0.081	& 0.822	& 0.504\\
ComplEx~\cite{Trouillon:ComplEx} & 0.401	& 0.642	& 0.124	& 0.377	 & 0.396	& 0.086	& 0.839	& 0.529\\
RotatE~\cite{RotatE}	  & 0.552	& 0.664	& 0.138	& 0.404	 & 0.552	& 0.111	& \underline{0.872}	& 0.553\\
PairRE~\cite{PairRE}	& 0.563	& 0.651	& 0.139	& 0.407	 & 0.542	& 0.098	& 0.871	& 0.562\\
HAKE~\cite{HAKE}	& 0.573	& 0.657	& 0.162	& 0.411	 & 0.568	& 0.116	& 0.862	& 0.550\\
\midrule
HAKE+ECSE	& \underline{0.583}	& \underline{0.673}	& \underline{0.191}	& \underline{0.441}	 & \underline{0.573}	& \bf{0.147}	& \bf{0.875}	& \underline{0.596}\\
HAKE+ICSE	& \bf{0.604}	& \bf{0.675}	& \bf{0.213}	& \bf{0.463}	 & \bf{0.578}	& \underline{0.141}	& \bf{0.875}	& \bf{0.600}\\
\bottomrule
\end{tabular}
\label{table7}
\end{table*}

We compare our common sense-guided negative sampling (CGNS) mechanism with various categories of negative sampling (NS) techniques, including uniform sampling (Unifo)~\cite{Bordes:TransE}, none sampling (NoSamp)~\cite{nonesampling}, NSCaching (NSCach)~\cite{zhang2019nscaching}, domain-based sampling (DomSam)~\cite{DomainSampling} and self-adversarial sampling (SAdv)~\cite{RotatE}. The comparison results are obtained by combining these NS techniques$\footnote{Uniform sampling: \url{https://github.com/thunlp/KB2E}, none sampling: \url{https://github.com/rutgerswiselab/NS-KGE}, NSCaching: \url{https://github.com/yzhangee/NSCaching}, 
self-adversarial sampling: \url{https://github.com/DeepGraphLearning/KnowledgeGraphEmbedding}.}$ with the most classical KGE model TransE\cite{Bordes:TransE}. From the results shown in Table \ref{table5}, our CGNS significantly outperforms all the other NS techniques. Specifically, domain-based NS, self-adversarial sampling and our CGNS mechanism perform better than the others via the high-quality negative triples. Furthermore, our CGNS module performs the best benefited from its ability of avoiding false-negative triples and further improving the quality of negative triple by common sense. These results illustrate the superiority of our CGNS strategy to generate more high-quality negative triples for KGE models.

\subsubsection{\textbf{Results of Our Framework with Implicit Entity Concepts}}

The results of our ICSE model integrated into some basic KGE models are exhibited in Table~\ref{table6}. Our scheme outperforms each corresponding basic KGE model on all the metrics except for MR. Specific to the metric Hits@1 on datasets FB15K/FB15K237/WN18/YAGO3-10, our ICSE model achieves improvements of 10.9\%/27.3\%/23.9\%/37.9\% compared to TransE, 25.4\%/3.4\%/67.9\%/123.5\% compared to ComplEx, 4.7\%/27.4\%/0.7\%/29.5\% compared to RotatE, 1.6\%/3.2\%/4.9\%/- (HAKE cannot be conducted on YAGO3-10 due to limited computility) compared to HAKE. Thus, scoring triple candidates in a joint common sense and fact fashion by virtue of relation-aware concept embeddings facilitates better performance of KGE models. These results illustrate the superiority and effectiveness of conveniently integrating implicit common sense to improve the original KGE models.

\subsubsection{\textbf{Results on Complex Properties of Relations}}

We conducted an evaluation with a focus on complex relations specific to datasets FB15K and FB15K237 for more diverse relations in these two datasets compared to others. As shown in Table~\ref{table7}, our framework consistently outperforms other baselines in both head entity prediction and tail entity prediction. Notably, it excels in the most challenging tasks, namely predicting head entities on N-1 relations and predicting tail entities on 1-N relations. Besides, ECSE performs better than ICSE. These results demonstrate that our framework especially the explicit common sense facilitates KGC on complex properties of relations more effectively.

\subsection{Ablation Study}

To demonstrate the effectiveness of each contribution of ECSE, we construct the following ablated models integrated with the basic model HAKE:
\begin{itemize}
    \item -CRNS: this ablated model neglects the complex relation-aware sampling in common sense-guided negative sampling to generate negative triples.
    \item -CSNS: we remove the common sense-enhanced sampling from our negative sampling. This signifies that all entities can be exploited to construct negative triple candidates without the common sense constraint.
    \item -CFI: this ablated model omits the coarse-grained concept filtering in the view of common sense, leading to the ordinary inference manner of the existing KGE models.
\end{itemize}

\begin{figure*}
    \centering
    \includegraphics[width=1.0\textwidth]{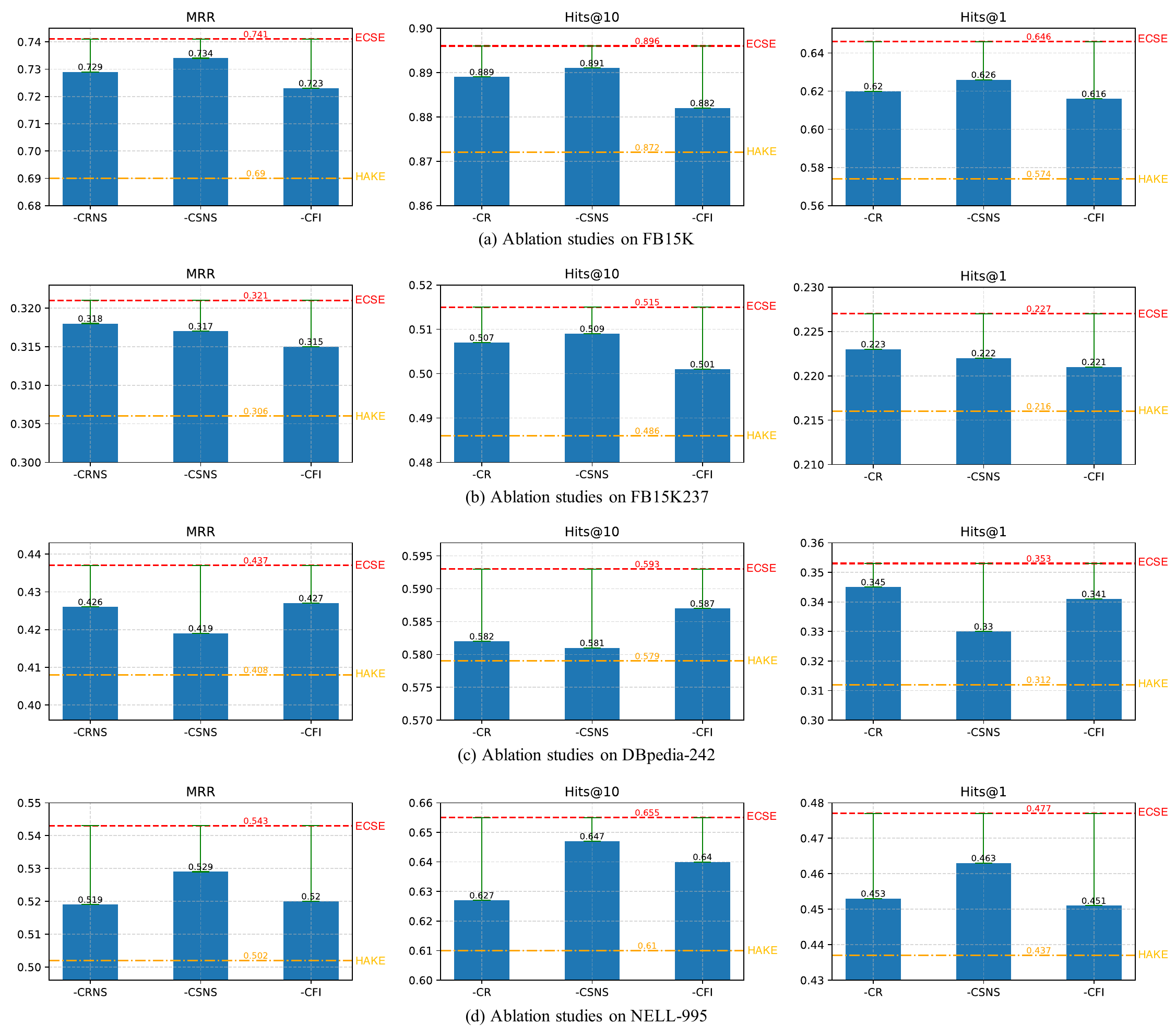}
    \caption{Ablation studies of ECSE model on four datasets. The red line represents HAKE+ECSE model, while the yellow line indicates HAKE model.}
    \label{fig:ablation_ECSE}
\end{figure*}

The results as shown in Fig.~\ref{fig:ablation_ECSE}, demonstrate that our complete model ECSE consistently outperforms all the ablated models on each dataset. This emphasizes the effectiveness of common sense-enhanced sampling and complex relation-aware sampling for generating high-quality negative triples. Furthermore, the coarse-grained concept filtering module plays a crucial role in enhancing KGC performance based on prior common sense. In summary, each contribution within the ECSE framework is pivotal in achieving these improvements.

To verify the performance of each module within ICSE model, the ablated models are designed as followings:

\begin{itemize}
    \item -ENT: we wipe out entity embeddings together with factual triples to access the inference results in the single view of common sense.
    \item -ConE: this ablated model removes concept embeddings and implicit common sense triples from ICSE model. It conducts KGC merely relying on factual triples.
    \item -CESR: in this ablation model, the concept embedding similarity regularization is neglected when learning concept embeddings in the training procedure.
\end{itemize}

\begin{figure*}
    \centering
    \includegraphics[width=1.0\textwidth]{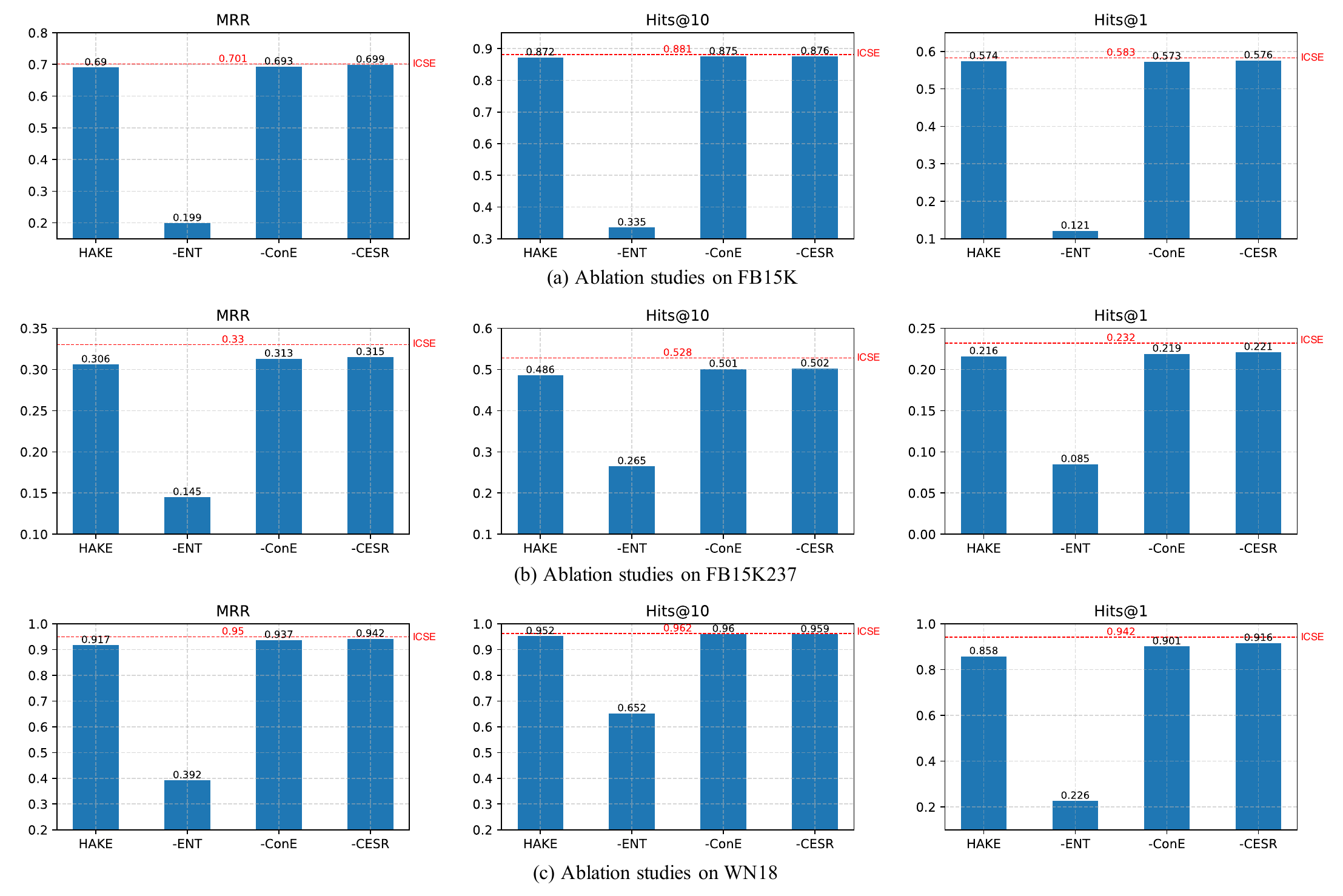}
    \caption{Ablation studies specific to the ICSE model on three datasets. The red dashed line represents the performance of our complete HAKE+ICSE model.}
    \label{fig:ablation_ICSE}
\end{figure*}

The comparison results of the ablated models and the whole model ICSE as well as the original basic model HAKE are shown in Fig.~\ref{fig:ablation_ICSE}. We reveal several key insights:

(1) According to the results of -ConE, representing implicit common sense triples with concept embeddings contributes to higher precision in KGC tasks by assessing the plausibility of triple candidates from a common sense perspective.

(2) Concept embedding similarity regularization leads to more effective representation of concepts in the context of the associated relation.

(3) Notably, -ENT model exhibits more performance drop compared to other models. This highlights that evaluating the plausibility of a triple candidate based solely on implicit common sense triples is insufficient for capturing the individual features of factual triples for the entity-specific KGC tasks.

\subsection{Case Study}

To demonstrate the superiority of our framework compared with the classical models ignoring common sense, we exhibit the case study results by HAKE+ECSE/ICSE and the baseline model HAKE on NELL-995. From the inference results of predicting missing head or tail entities shown in Table~\ref{table11}, our models HAKE+ECSE and HAKE+ICSE both achieve the higher rank of the correct entity (rank 1) for each test instance compared to the baseline model HAKE, which emphasizes the superiority of introducing common sense into KGE model to improve the precision of KGC.

\begin{table}
  \caption{Some cases of the KGC results obtained by HAKE+ECSE and HAKE+ICSE as well as baseline model HAKE on NELL-995.}
  \renewcommand{\arraystretch}{0.6}
  \begin{tabular}{l|l}
  \toprule
  \multicolumn{2}{c}{Case 1: $(soledad\_o\_brien, worksfor, ?)$}\\
  \midrule
    HAKE    & \textbf{rank} of the correct tail entity $cnn$: 11 \\
    \midrule
                & \textbf{rank} of the correct tail entity $cnn$: 1 \\
    HAKE+ECSE   & common sense: $(journalist, worksfor, $ \\
                & $\{sportsteam, company, organization, ...\})$ \\
                & concept of the correct entity: $company$ \\
    \midrule
    HAKE+ICSE   & \textbf{rank} of the correct tail entity $cnn$: 1 \\
  \bottomrule
  \toprule
  \multicolumn{2}{c}{Case 2: $(jim\_rice, athleteplaysforteam, ?)$}\\
  \midrule
    HAKE    & \textbf{rank} of the correct tail entity $red\_sox$: 337 \\
    \midrule
                & \textbf{rank} of the correct tail entity $red\_sox$: 1 \\
    HAKE+ECSE   & common sense: $(athlete, athleteplaysforteam, $ \\ 
                & $\{sportsteam, geopoliticalloction\})$ \\
                & concept of the correct entity: $sportsteam$ \\
    \midrule
    HAKE+ICSE   & \textbf{rank} of the correct tail entity $red\_sox$: 1 \\
  \bottomrule
  \end{tabular}
  \label{table11}
\end{table}

In specific to Case 2, given a test instance with tail entity missing $(jim\_rice, athleteplaysforteam, ?)$, the rank of the correct entity $red\_sox$ achieved by HAKE is 337, signifying the challenge of predicting the tail entity caused by the uncertainty in embeddings relying solely on fact. In contrast, our models HAKE+ECSE and HAKE+ICSE both deduce the correct tail entity $red\_sox$ as the rank-1 candidate entity. This significant performance gain is obviously benefited from the explicit and implicit common sense. Furthermore, we can observe that the correct entity's concept $sportsteam$ conforms to the common sense $(athlete, athleteplaysforteam,$ $\{sportsteam, geopoliticalloction\})$ corresponding to this test instance. More interestingly, common sense and entity concepts not only compensate for the uncertainty in embeddings but also enhance the explainability of predicted results.

\section{Conclusion}

This paper proposes a pluggable common sense-enhanced KGC framework for improving the performance and scalability of KGC models. On account the KGs with rich entity concepts, we generate and take advantage of explicit common sense for improving KGE models via the common sense-guided negative sampling and a coarse-to-fine inference strategy. Specific to the KG lacking sufficient entity concepts, each factual triple is extended into the corresponding implicit common sense triple which can be represented by a relation-aware concept embedding mechanism. Furthermore, a dual score function is exploited to evaluate the plausibility of candidate triples from both fact and common sense perspectives. Extensive experimental results illustrate that our framework achieve significant and consistent performance improvements on KGC tasks via introducing common sense into KGE models. In the future, we plan to extend our framework by common sense derived from large language model.

\bibliographystyle{IEEEtran}
\bibliography{IEEEabrv}

\begin{IEEEbiography}[{\includegraphics[width=1in,height=1.25in,clip,keepaspectratio]{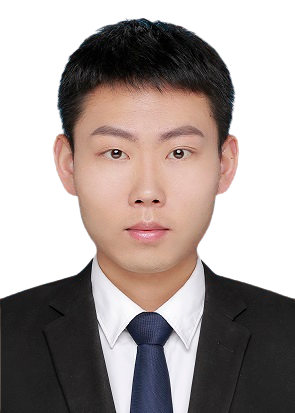}}]{Guanglin Niu}
received the bachelor's degree in automation from Beijing Institute of Technology in 2015, and the Ph.D. degree in computer science from Beihang University in 2022. He is currently a assistant professor with the School of Artificial Intelligence, Beihang University. He has authored some top-tier conference papers as first author, including AAAI, ACL, SIGIR, EMNLP and COLING. His research interests include machine learning, knowledge graph and natural language processing.
\end{IEEEbiography}

\begin{IEEEbiography}[{\includegraphics[width=1in,height=1.25in,clip,keepaspectratio]{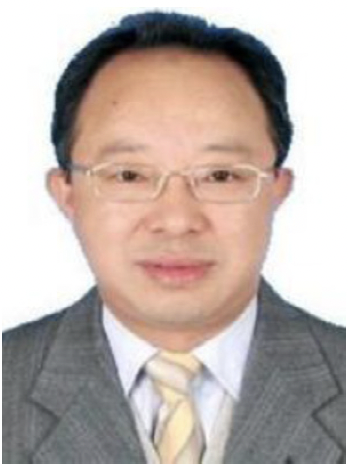}}]{Bo Li}
is currently a Changjiang Distinguished Professor with the School of Artificial Intelligence, Beihang University. He is a recipient of The National Science Fund for Distinguished Young Scholars. He is currently the associate dean of the School of Artificial Intelligence, Beihang University. He is the principal investigator of the National Key Research and Development Program. He has published over 100 papers in top journals and conferences and held over 50 domestic and foreign patents.
\end{IEEEbiography}

\begin{IEEEbiography}[{\includegraphics[width=1in,height=1.25in,clip,keepaspectratio]{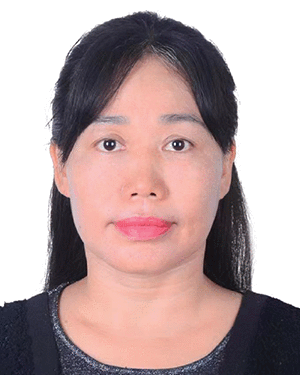}}]{Siling Feng}
received the Ph.D. degree from the University of Electronic Science and Technology of China in 2014. She is currently a Professor and a Ph.D. Supervisor with the School of Information and Communication Engineering, Hainan University, Haikou, China. Her research interests include intelligent computing, Big Data analysis, and intelligent recommendation.
\end{IEEEbiography}

\vfill

\end{document}